\documentclass[10pt,twocolumn,letterpaper]{article}

\usepackage[pagenumbers]{cvpr} 

\usepackage{multirow}
\usepackage{tabularx}
\usepackage{float}
\usepackage{caption}
\usepackage{graphicx}%
\usepackage{multirow}%
\usepackage{amsmath,amssymb,amsfonts}%
\usepackage{amsthm}%
\usepackage{mathrsfs}%
\usepackage[title]{appendix}%
\usepackage{textcomp}%
\usepackage{manyfoot}%
\usepackage{booktabs}%
\usepackage{algorithm}%
\usepackage{algorithmicx}%
\usepackage{algpseudocode}%
\usepackage{listings}%
\usepackage{multicol}
\usepackage{makecell}
\usepackage{multirow}
\usepackage{caption}
\usepackage{bm}
\definecolor{darkred}{rgb}{0.55, 0.0, 0.0}  
\definecolor{darkgreen}{rgb}{0.0, 0.5, 0.0}  
\definecolor{cvprblue}{rgb}{0.21,0.49,0.74}
\usepackage[pagebackref,breaklinks,colorlinks,allcolors=cvprblue]{hyperref}
\usepackage{booktabs}
\def\paperID{7975} 
\def\confName{CVPR}
\def\confYear{2025}

\title{FlashSloth \includegraphics[width=0.9cm]{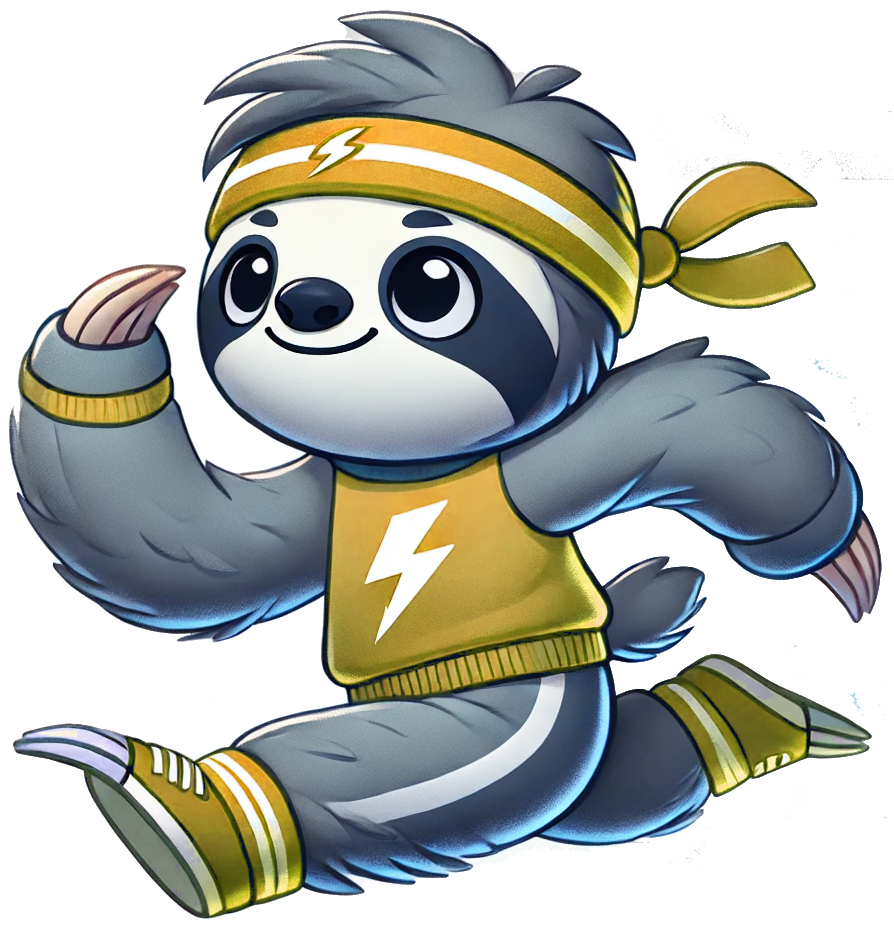}: Lightning Multimodal Large Language Models via \\ Embedded Visual Compression }

\author{
Bo Tong$^{1}$, Bokai Lai$^{1}$, Yiyi Zhou$^{1}$\thanks{Corresponding Author.}, Gen Luo$^{3}$, Yunhang Shen$^{2}$, Ke Li$^{2}$,  Xiaoshuai Sun$^{1}$, Rongrong Ji$^{1}$\\
$^{1}$ Key Laboratory of Multimedia Trusted Perception and Efficient Computing, \\
Ministry of Education of China, Xiamen University, 361005, P.R. China. \\
$^{2}$ Youtu Lab, Tencent, P.R. China.\\
$^{3}$ OpenGVLab, Shanghai AI Laboratory.\\
{\tt\small \{tongbo,laibokai\}@stu.xmu.edu.cn, \{zhouyiyi, xssun, rrji\}@xmu.edu.cn,}
\\
{\tt\small luogen@pjlab.org.cn, shenyunhang01@gmail.com, tristanli@tencent.com}
}

\begin{document}
\maketitle
\begin{abstract}
Despite a big leap forward in capability, \emph{multimodal large language models} (MLLMs) tend to behave like a sloth in practical use, \emph{i.e.}, slow response and large latency. Recent efforts are devoted to building tiny MLLMs for better efficiency, but the plethora of visual tokens still used limit their actual speedup. In this paper, we propose a powerful and fast tiny MLLM called \emph{\textbf{FlashSloth}}. Different from previous efforts, FlashSloth focuses on improving the descriptive power of visual tokens in the process of compressing their redundant semantics. In particular, FlashSloth introduces embedded visual compression designs to capture both visually salient and instruction-related image information, so as to achieving superior multimodal performance with fewer visual tokens. Extensive experiments are conducted to validate the proposed FlashSloth, and a bunch of tiny but strong MLLMs are also comprehensively compared, e.g., InternVL2, MiniCPM-V2 and Qwen2-VL. The experimental results show that compared with these advanced tiny MLLMs, our FlashSloth can greatly reduce the number of visual tokens, training memory and computation complexity while retaining high performance on various VL tasks. Our code is released at: \url{https://github.com/codefanw/FlashSloth}.
\end{abstract}
    
\section{Introduction}
\label{sec:intro}

Recent years have witnessed the remarkable breakthroughs made by extending \emph{large language models} (LLMs)~\cite{llama, phi, opt} to more modalities, e.g., building \emph{multimodal large language models} (MLLMs) for vision-language tasks ~\cite{blip, llava, lavin}. Among these advancements, one main research focus is on enhancing the  visual perception of MLLMs, and the widely recognized solution is to use a larger number of visual tokens ~\cite{llavanext, monkey, llavahr}. For instance, LLaVA-NeXT ~\cite{llavanext} uses 5 times more visual tokens compared to LLaVA-1.5~\cite{llava1.5} by subdividing input images into multiple tiles. Similarly, recent MLLMs, such as InternVL1.5 ~\cite{internvl} and Qwen2-VL~\cite{wang2024qwen2}, can support up to thousands of visual tokens for high-resolution image understanding via dynamic-resolution encoding. Although effective, the excessive use of visual tokens further execrates already high computation of MLLMs, limiting practical use. 

\begin{figure}[t]
  \centering
   \includegraphics[width=\linewidth]{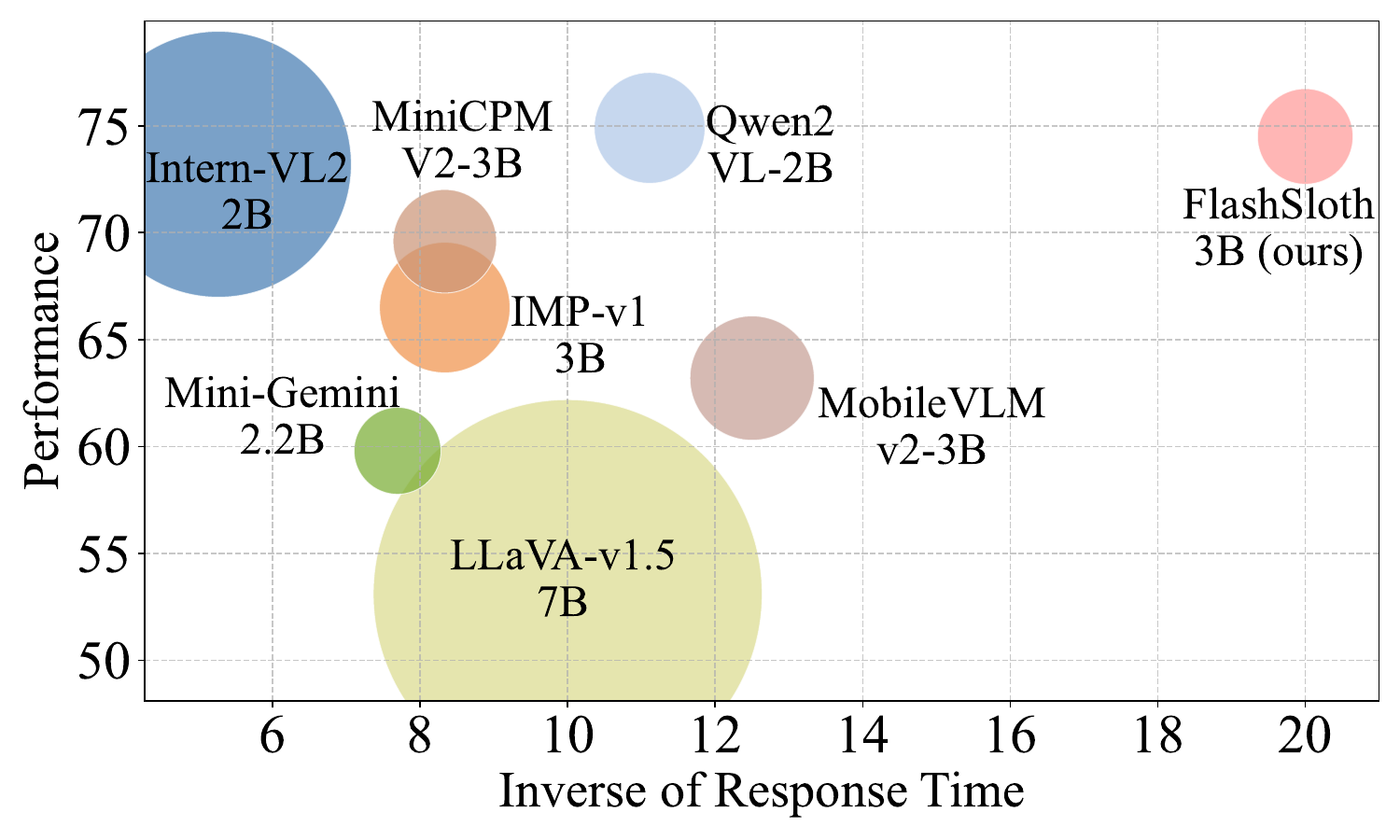}

   \caption{Comparison between FlashSloth and recent MLLMs on MMB in terms of performance, response time (the prediction of first token) and GPU memory overhead (the circle size). Advanced tiny MLLMs~\cite{minigemini, chu2024mobilevlm, shao2024imp, minicpm, wang2024qwen2,internvl} can already exhibit strong capability against common MLLMs like LLaVA-1.5-7B~\cite{llava1.5}, but their actual speed up is greatly limited by the excessive use of visual tokens. Our FlashSloth is a powerful and tiny MLLM that offers a decent balance between performance and efficiency.}
   \label{fig:1}
   \vspace{-0.3cm}
\end{figure}

In this case, more and more efforts are devoted to the research of lightweight and efficient MLLMs  ~\cite{chu2024mobilevlm, shao2024imp, hu2024minicpmunveilingpotentialsmall, yuan2024tinygptvefficientmultimodallarge}. In particular, with the emergence of small-scale LLMs, \emph{e.g.}, Phi~\cite{phi}and Gemma~\cite{gemma}, recent endeavors start to explore their use in building tiny MLLMs, such as MobileVLM~\cite{chu2023mobilevlm, chu2024mobilevlm}, Imp~\cite{shao2024imp} and Mini-Gemini~\cite{minigemini}. Meanwhile, representative MLLM families also launch their slim versions for better mobile applications, \emph{e.g.}, Qwen2-VL~\cite{wang2024qwen2}, InternVL~\cite{internvl} and MiniCPM-V~\cite{hu2024minicpmunveilingpotentialsmall}. With a much smaller LLM structure, these tiny MLLMs typically scale to about 2-3 billion parameters, so their training expenditure as well as memory overhead are also much cheaper than previous MLLMs~\cite{llava1.5, blip, lavin}. However, to retain general multimodal capability, most tiny MLLMs~\cite{shao2024imp, yuan2024tinygptvefficientmultimodallarge, internvl} still adopt a large number of visual tokens, making it hard to achieve actual speedup. As shown in Fig.~\ref{fig:1}, with more visual tokens used, tiny MLLMs even have a slower response time\footnote{The time for the first answer token.} than common MLLMs like LLaVA-1.5-7B~\cite{llava1.5}.

By revisiting the development of vision-language research~\cite{show,trar,stacked,bottom,luo1}, we can see that the way to achieve better visual capability is not confined to a singular paradigm. In principle, the key to addressing visual shortcoming is to make ``\emph{vision}'' matter in MLLMs~\cite{making, need}, thereby helping them better understand visual information and also reduce the impact of \emph{language bias}~\cite{free, contrast}. From this perspective, the use of enough visual tokens does contribute more to self-attention modeling in MLLMs, but recent studies~\cite{fastv, ye2024fit} also show that this paradigm is often inefficient and obviously redundant. In addition, various attempts have been successfully made in improving visual capability before the era of MLLMs. For instance, enriching the visual semantics~\cite{in, vinvl,luo2} or refining complex image information based on visual saliency or question dependency through various attention-based approaches~\cite{deep, bottom, trar, luo3}. To this end, we believe that a good balance between the performance and efficiency of MLLMs is feasible. 

In this paper, we propose a tiny and fast MLLM called \emph{FlashSloth}. The principle of FlashSloth is to improve the descriptive power of visual tokens in the process of refining and compressing their redundant semantics, thereby achieving actual speedup during inference. Concretely, FlashSloth first introduces a \emph{spatial-aware attentive pooling} to compress the redundant image information while capturing visually salient semantics. Meanwhile, a novel and lightweight \emph{Query} module is equipped to grasp  instruction-related image information, thereby compensating the loss of image details in the attention pooling process. Notably, this query module is embedded into the architecture of FlashSloth rather than as an independent bridge branch that requires another language modeling~\cite{minigpt, instructblip, mplug}, e.g., Q-Former~\cite{blip}. Thus, we term it \emph{EmbQ}. In addition to the compact structure designs, EmbQ also consumes much lower training and inference costs, well facilitating the efficiency goal of FlashSloth. For instance, EmbQ does not require dedicated large-scale VL alignment pretraining~\cite{blip}. With these intuitive designs, FlashSloth can not only greatly reduce the number of input visual tokens but also improve their discrimination for better multimodal reasoning.

To validate the proposed FlashSloth, we conduct extensive experiments on a set of highly-competitive VL and MLLM benchmarks~\cite{mmb,mme,mmmu,seed,pope}, and compare it with a bunch of least tiny MLLMs, including Qwen2-VL-2B~\cite{wang2024qwen2}, Intern-VL-2~\cite{internvl}, MiniCPM-V2~\cite{minicpm}, and MM1.5~\cite{mm1}. Experimental results demonstrate that compared to these advanced tiny MLLMs, our FlashSloth can reduce the number of visual tokens, training memory and inference computation by 80-89\%, 61-80\% and 70-98\%, respectively, while shortening the actual response time by about 2$\times$ to 5$\times$ times. Retaining high efficiency, FlashSloth also exhibits competitive ness against these SOTA methods, and even perform slightly better on several common VL tasks, \emph{e.g.}, MMB~\cite{mmb} and MMMU~\cite{mmmu}, well confirming our motivation and the designs of FlashSloth.
\\In summary, our contributions are three folds:

\begin{itemize}

     \item We propose a strong and fast tiny MLLM in this paper, coined as \emph{FlashSloth}, showing that a good balance between performance and efficiency is feasible.  
     
     \item In FlashSloth, we introduce embedded visual compression designs to efficiently capture both visually salient and instruction-related semantics, namely \emph{Spatial Attention Pooling} and \emph{Embedded Query} modules.
     
     \item The extensive experiments not only show the strong multimodal capability of FlashSloth, but also confirm its competitiveness with a set of advanced MLLMs while retaining higher efficiency.
 \end{itemize}
\begin{figure*}
  \centering
    \includegraphics[width=\linewidth]{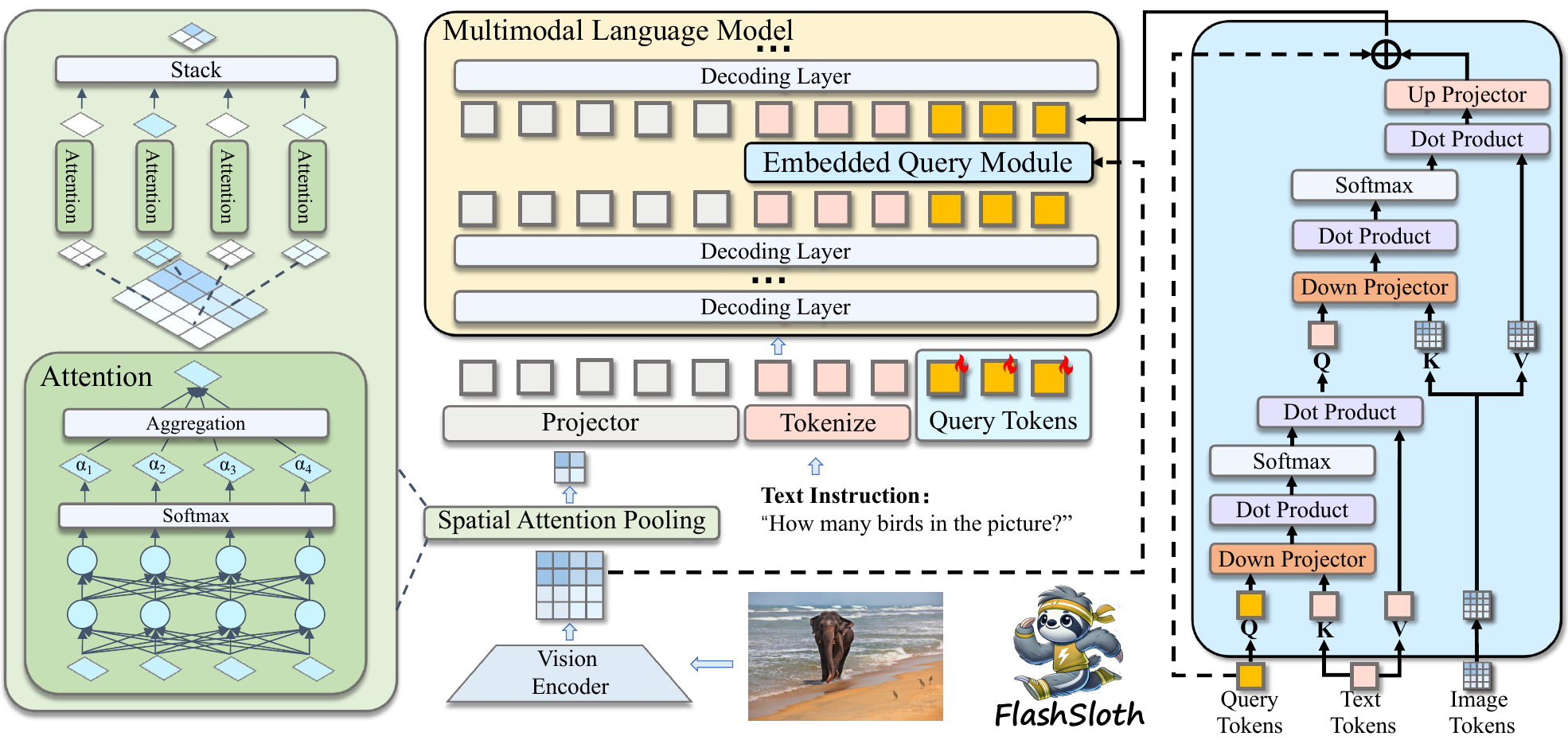}
    \caption{The overall framework of the proposed FlashSloth. The visual tokens extracted by the vision encoder are first refined and compressed by a \emph{Spatial Attention Pooling} (SAP) module and then fed to FlashSloth. In addition to visual and text tokens, a set of learnable query tokens are also padded to query instruction-related image information via the \emph{Embedded Query Module} (EmbQ) after some layers of FlashSloth. In particular, SAP is to capture the visually salient semantics in image regions via uni-modal visual attention, as depicted in the left. EmbQ is a lightweight and embedded module for visual enhancement in FlashSloth, which requires no additional language modeling and dedicated alignment pretraining, as shown in the right.}
    \label{fig:2}
    \vspace{-0.3cm}
\end{figure*}

\section{Related Works}

\subsection{Multimodal Large Language Models}

Based on the rapid development of \emph{large language models} (LLMs) \cite{llama,opt,phi} and visual encoders \cite{radford2021learning, oquab2023dinov2}, \emph{multimodal large language models} (MLLMs) also achieve  significant strides in various \emph{vision-language} (VL) tasks. Numerous open-source MLLMs~\cite{li2022blip, li2023blip, lavin, llava1.5}  emerges in recent years, some of which even achieve outstanding capability comparable to GPT-4~\cite{gpt4} in specific fields. However, this advancement is often support with increasingly larger parameter sizes, which also results in heavy burden to the training and application of MLLMs. Therefore, more recent research resorts to smaller LLMs to build tiny MLLMs, such as Phi~\cite{phi}, Gemma~\cite{gemma}, and Qwen2~\cite{bai2023Qwen}. For instance, MobileVLM~\cite{chu2023mobilevlm} first realize the attempt of extending tiny LLMs to multimodal tasks with a simple yet effective visual projection after image encoder. Additionally,  more efforts are devoted to explore the design and training strategies of tiny MLLMs based on small LLMs, such as LLaVA-Phi~\cite{llavaphi}, Imp~\cite{shao2024imp}, and PaliGemma~\cite{chen2022pali}. Meanwhile, the influential MLLM families, such as MiniCPM-V~\cite{minicpm}, Qwen2-VL~\cite{wang2024qwen2} and InternVL~\cite{internvl}, also develop their slim but also powerful versions via exploring high-resolution image encoding and high-quality data collection. Overall, the advancement of tiny MLLMs well facilitate the real-world applications of MLLMs. However, a magnitude of visual tokens still used also slow down the response time of tiny MLLMs in addition to high expenditure~\cite{tokenpacker, fastv}, \emph{i.e.}, the first token prediction, hindering application.
 
\subsection{Visual Token Compression}
Most existing MLLMs~\cite{eagle,wang2024qwen2,minigemini,internvl,llavahr} usually rely on a large number of visual tokens for superior visual capability, whether they are large or tiny. However, this paradigm is often criticized for excessive computation and obvious visual redundancy, which also attracts an influx of interest in efficient visual learning of MLLMs~\cite{fastv,ye2024fit,tome}.
In terms of network designs, \emph{Q-Former-}like methods~\cite{blip, instructblip, minigpt, mplug} uses learnable tokens to control the number of visual tokens, with a purpose of capturing instruction-related visual information via visual compression. However, they often use another language model like BERT~\cite{bert} to interpret the text instruction and require dedicated vision-language pretraining. Some methods like  Abstractor~\cite{honeybee} and LDP~\cite{chu2023mobilevlm,chu2024mobilevlm} employ convolutional layers to learn local visual compression. Similarly, methods \cite{internvl} like InternVL apply pixel shuffle to directly reduce the number of visual tokens. However, the information loss in these local compression methods are often not further compensated. The other main paradigm for visual efficiency is to apply external methods for effective visual token pruning during inference \cite{fastv, ye2024fit, tome}. For example, FastV~\cite{fastv} determines each token’s importance based on average attention,and FitPrune~\cite{ye2024fit} selects retained features by minimizing the attention distribution difference before and after pruning. However, the contributions of token pruning methods are orthogonal to this paper, and we focus on improving the discrimination of visual tokens via investigating network structure design.
\section{Method}
\subsection{Overall}
\label{Overall}
In this paper, we propose a tiny and fast MLLM called \emph{FlashSloth}, of which framework is illustrated in Fig.~\ref{fig:1}. FlashSloth aims to improve the descriptive power of visual tokens with embedded visual compressions, \emph{i.e.}, the \emph{spatial attention pooling} (SAP) and \emph{embedded query } (EmbQ) modules, thereby achieving superior multimodal capability with a shorter visual token sequence.

Concretely, given an input image $I$, FlashSloth uses the image encoder to extract its visual token features, denoted $\mathbf{F}_v\in\mathbb{R}^{(h\times w)\times d}$, where $h\times w$ denotes the resolution and $d$ is the feature dimension. And the input text instruction $T$ is first truncated into a set of tokens, which are then vectorized by the corresponding word embeddings, denoted as $\mathbf{F}_t\in\mathbb{R}^{l \times d}$. Here, $l$ is the length of text sequence. 

In existing MLLMs~\cite{llava1.5, shao2024imp}, the number of directly output visual tokens $\mathbf{F}_v$ is often large, especially for the high-resolution images~\cite{llavanext,internvl,minicpm}. Thus, we apply the SAP module to attentively capture the salient visual semantics while compressing the token redundancy. The processed visual tokens by SAP are denoted by $\mathbf{F}_v^s \in \mathbb{R}^{(\frac{w \times h}{s^2}) \times d}$, which has a much smaller number of tokens than $\mathbf{F}_v$. 

Afterwards, the compact visual tokens $\mathbf{F}_v^s$ after linear projection and the text tokens $\mathbf{F}_t$ are fed to the MLLM structure, which are also padded with $n$ learnable query tokens at the end of the sequence, denoted as $\mathbf{F}_q\in\mathbb{R}^{n\times d}$.

In FlashSloth, $\mathbf{F}_q$ serves to supplement $\mathbf{F}_v^s$ in terms of the instruction-related image details. Particularly, to avoid another language modeling~\cite{instructblip,minigpt,mplug}, $\mathbf{F}_q$ will first attend the multimodal interaction in the MLLM, and then engage in EmbQ for visual querying at the $k$-th layer:
\begin{equation}
\mathbf{F}_q^{(k)}= \mathbf{F}_q^{(k)}+ EmbQ(\mathbf{F}_v, \mathbf{F}_t^{(k)},\mathbf{F}_q^{(k)} ).
\end{equation}

Here, the output of EmbQ is pure visual attention features $\mathbf{F}_v^q\in \mathbb{R}^{n\times d}$ with the same length of $\mathbf{F}_q$. Overall, the objective of FlashSloth's decoding can be defined by:
\begin{equation}
p\left(A \mid \mathbf{F}_{\mathrm{v}}^s, \mathbf{F}_{\mathrm{t}},\mathbf{F}_{\mathrm{q}}\right)=\prod_{i=1}^{L} p\left(a_{i} \mid \mathbf{F}_{\mathrm{v}}^s, \mathbf{F}_{\mathrm{t}},\mathbf{F}_{\mathrm{q}}, \mathbf{F}_{a<i}\right),
\end{equation}
where $p$ is the prediction probability distribution, $A=\{a_1,...,a_L\}$ is the answer sequence and $L$ is its length. $\mathbf{F}_{a<i}$ denotes the answer tokens before the $i$-th step.

From the above introduction, we can see that FlashSloth is different from most MLLMs~\cite{llava1.5, internvl, wang2024qwen2} in two main aspects. First, FlashSloth refine and compress visual tokens in terms of both visual saliency and instruction-related semantics, which can well collaborate with each other for different VL tasks. Second, all compression designs are lightweight and embedded in the architecture of FlashSloth without the requirement of specific tuning or pretraining~\cite{blip}. In the following section, we will describe them in detail.
\subsection{Embedded Visual Compression}
As discussed above, the main principle of FlashSloth is to improve the discrimination of visual tokens while squeezing their length. To approach this target, we perform the visual compression in two aspects, \emph{i.e.,} \emph{visual saliency} and \emph{instruction dependency}. Moreover, these designs are lightweight and can be embedded into the architecture of FlashSloth, serving the target of model efficiency.
\subsubsection{Visually Salient Compression}
We first introduce a \emph{spatial attention pooling} (SAP) method to refine and compress the semantics in local visual tokens, borrowing the successful attention designs in previous VL research~\cite{trar}. The intuition is that the visual tokens extracted by the encoders like ViT~\cite{siglip, clip} already have a large receptive field as well as obviously overlapping information. Thus, an MLLM can mainly focus on the most visually salient information in each image regions.

Specifically, given the extracted visual tokens \( \mathbf{F}_v \in \mathbb{R}^{(h \times w) \times d} \), we first spatially divide them into a set of region tokens with a size of $s\times s$, denoted $\mathbf{F}_v^{i }  \in \mathbb{R}^{s^2 \times d}$. Thus, for each image region, SAP directly use a two-layer $mlp(\cdot)$ to predict its visual attention weights $\alpha \in \mathbb{R}^{s^2}$:
\begin{equation} \alpha = \text{Softmax}\big( mlp (\mathbf{F}_v^{i})\big).  \end{equation}
Then, the visually salient feature of each image region $f_{v}^s \in \mathbb{R}^d$ is directly obtained via weighed combinations:
\begin{equation} f_{v}^s = \sum_{i=1}^{s^2} \alpha^{i}\cdot f_{v}^i, \end{equation}
Lastly, those salient features are tiled to form the new visual tokens $\mathbf{F}_v^s \in \mathbb{R}^{(\frac{w \times h}{s^2}) \times d}$ and fed to FlashSloth.

\subsubsection{Instruction-aware Visual Compression}
Considering the varying difficulties of VL tasks~\cite{making,caption,rec}, the salient semantics provided by SAP is prone to insufficient for  multimodal reasoning. In this case, we further propose an \emph{embedded query} (EmbQ) module towards instruction-aware visual compression. 
In broad terms, EmbQ is similar to previous attempts like Q-Former~\cite{blip}, \emph{i.e.}, querying the text-related visual information, but it still exhibit obvious differences in design and operation. 

Above all, our requirement for EmbQ is to accomplish coarse-grained visual grounding rather than accurate VL alignment. By revisiting  previous VL research~\cite{show,stacked}, we note that this requirement is easy to achieved without complex network structure and large-scale pretraining. Therefore, the design of EmbQ is neat and efficient, which is directly embedded into FlashSloth, as shown in Fig.~\ref{fig:2}. 

Concretely, a set of learnable tokens $\mathbf{F}_q\in \mathbb{R}^{n \times d} $ are used as \emph{queries} and padded in the input sequence of FlashSloth. After $k$ layers of transformation, $\mathbf{F}_q^{(k)}$ are fed to EmbQ for visual querying. In particular, we expect this operation will allow $\mathbf{F}_q$ to obtain enough instruction information from the text tokens via self-attention. But during experiments, we note that more visual semantics are received since the length of visual tokens is much longer than that of the text ones, which contradicts the target of EmbQ. 

Thus, we first interact  $\mathbf{F}_q^{(k)}$ and $\mathbf{F}_t^{(k)}$ via cross-attention:
 \begin{equation}  \mathbf{F}_t^q = \text{Softmax} \left( \frac{(\mathbf{F}_{q}^{(k)}\mathbf{W}_q ) ( \mathbf{F}_{t}^{(k)}\mathbf{W}_k)^T}{\sqrt{d_k}} \right) \mathbf{F}_{t}^{(k)}\mathbf{W}_v . \label{t2q}\end{equation}
where $\mathbf{F}_t^q \in \mathbb{R}^{n \times d}$ are the obtained text queries, and $\mathbf{W}_q, \mathbf{W}_k, \mathbf{W_v} $ are the projection weight matrices, and $d_k$ denotes their dimension. 

Then, we can use $\mathbf{F}_t^q$ to query visual information from the uncompressed visual tokens $\mathbf{F}_v$, defined by
\begin{equation} \mathbf{F}_{v}^q = \text{Softmax} \left( \frac{(\mathbf{F}_t^q\mathbf{W}_q ) (\mathbf{F}_{v}\mathbf{W}_k)^T}{\sqrt{d_k}} \right) \mathbf{F}_{v}\mathbf{W}_v . \label{i2q}\end{equation}

Lastly,  \( \mathbf{F}_{v}^q \) are up projected and then combined with $\mathbf{F}_q^{(k)}$  for the following multimodal inference of FlashSloth, as described in Sec. \ref{Overall}. Notably, the process of EmbQ takes into account of the discrimination of well-learned visual tokens, so only one up-projection is used for visual tokens for scaling to the MLLM's dimension. Besides, we also use the embedding of ‘\emph{dot}’ token to initialize queries, making them easier to accommodate to the semantic space of MLLM.

\subsection{Training and Other Settings}
Under the default setting, FlashSloth apply a two-stage training paradigm~\cite{llava}.\\
\textbf{The pretraining stage.} Only the projector and spatial attention pooling are optimized for the alignment between visual and text tokens, while the LLM is fixed.\\
\textbf{The SFT tuning stage.} Except vision encoder, FlashSloth are optimized, including LLM and EmbQ. 

To tackle OCR tasks with a high demand on image resolution~\cite{llavanext,llavaonevision,internvl}, we also propose a high-resolution version, termed \textbf{FlashSloth-HD}. In particular, FlashSloth-HD inputs images of $768\times 768$ resolutions. In terms of image processing, we follow LLaVA-NeXT \cite{llavanext} to divide the images into four parts and a low-resolution thumbnail, of which visual tokens are extracted in parallel.  Similarly, FlashSloth use SAP to squeeze their length greatly and EmbQ for visual querying. To save training expenditure, we only use high-resolution images in the SFT tuning of FlashSloth-HD, where the vision encoder is unfreezed for better accommodation. Details can refer to our project.

\section{Experiment}
\subsection{Implementation Details}
In terms of the default FlashSloth, we use \emph{siglip-so400m}\cite{siglip} as the visual encoder with an input resolution of 384, and \emph{phi2-2.7B}\cite{phi} as the LLM. The downsampling rate of SAP is set to 3, generating 81 visual tokens. The number of queries in EmbQ is 9, and an embedding query module with a dimension of 576 is inserted at the 8th layer of the MLLM by default. The model is trained by \emph{AdamW}~\cite{adam} optimizer and cosine learning rate scheduler for a total of 1 epoch. The initial learning rates for pre-training and instruction tuning are 1e-3 and 2e-5 with batch sizes of 256 and 128, respectively. All training is conducted on 8 NVIDIA A800 GPUs. For the FlashSloth-HD, the input image resolution is 768, and the image tokens are compressed to 405, while the other settings remain the same as FlashSloth.
\subsection{Benchmarks and Metrics}
We evaluate the model on seven multimodal benchmark datasets, including MMB~\cite{mmb}, MME~\cite{mme}, mm-vet~\cite{mmvet}, Pope~\cite{pope}, SEED~\cite{seed}, MMMU~\cite{mmmu}, and MathVista~\cite{mathvista}. And seven general visual-language  datasets, including SQA~\cite{sqa}, AI2D~\cite{ai2d}, GQA~\cite{gqa}, TextVQA~\cite{textvqa}, ChartQA~\cite{chartqa}, DocVQA~\cite{docvqa}, and RealWorldQA. These benchmarks assess MLLMs from diverse perspectives, such as hallucination, multimodal perception and cognition, and multidisciplinary question answering. All evaluations are conducted using the \emph{lmms-eval}~\cite{lmms-eval}.
\begin{table*}[h]
\centering
\renewcommand{\arraystretch}{1.0}
\fontsize{8}{3.5}\selectfont
\setlength{\abovecaptionskip}{0.0cm}
\setlength{\belowcaptionskip}{0.3cm}
\caption{A comparison of FlashSloth with the latest MLLMs in terms of inference efficiency across five tasks. The comparison metrics include the number of visual tokens, first-round inference TFLOPs, inference memory usage (GB), model response time (seconds), and throughput (samples/second). The percentages in the table show the change in tiny MLLMs compared to LLaVA with red arrows indicating a decrease in performance and green arrows indicating an improvement. The best results are \textbf{bold} and the second-best results
are \underline{underlined}.
}
\resizebox{\textwidth}{!}{

\begin{tabular}{@{}l|l|c|ccccc@{}}
\toprule
Task                     & Metric        & \begin{tabular}[c]{@{}c@{}}LLaVA~\cite{llava1.5}\\V1.5-7B\end{tabular} & \begin{tabular}[c]{@{}c@{}}IMP~\cite{shao2024imp}\\ 3.1B\end{tabular}                                        & \begin{tabular}[c]{@{}c@{}}Qwen2-VL~\cite{wang2024qwen2}\\ 2B\end{tabular} & \begin{tabular}[c]{@{}c@{}}InternVL2~\cite{internvl}\\ 2B\end{tabular} & \begin{tabular}[c]{@{}c@{}}FlashSloth-HD\\ 3.2B\end{tabular} & \begin{tabular}[c]{@{}c@{}}FlashSloth\\ 3.2B\end{tabular} \\ \midrule
\multirow{5}{*}{GQA}     & Token number  & 576                                                     & 729\textsubscript{\textcolor{darkred}{$\uparrow 27\%$}} & \underline{352}\textsubscript{\textcolor{darkgreen}{$\downarrow 39\%$}}                                                   & 1699\textsubscript{\textcolor{darkred}{$\uparrow 195\%$}}                                                    & 414\textsubscript{\textcolor{darkgreen}{$\downarrow 28\%$}}                                                           & \textbf{90}\textsubscript{\textcolor{darkgreen}{$\downarrow 84\%$}}                                                        \\
                         & TFLOPs       & 4.41                                                    & \underline{0.30}\textsubscript{\textcolor{darkgreen}{$\downarrow 93\%$}}                                                                                      & 1.50\textsubscript{\textcolor{darkgreen}{$\downarrow 66\%$}}                                                  & 5.10\textsubscript{\textcolor{darkred}{$\uparrow 16\%$}}                                                    & 2.71\textsubscript{\textcolor{darkgreen}{$\downarrow 39\%$}}                                                          & \textbf{0.08}\textsubscript{\textcolor{darkgreen}{$\downarrow 98\%$}}                                                      \\
                         & GPU memory   & 15.3                                                    & 8.4\textsubscript{\textcolor{darkgreen}{$\downarrow 45\%$}}                                                                                       & 9.3\textsubscript{\textcolor{darkgreen}{$\downarrow 39\%$}}                                                   & 9.3\textsubscript{\textcolor{darkgreen}{$\downarrow 39\%$}}                                                     & \underline{7.8}\textsubscript{\textcolor{darkgreen}{$\downarrow 49\%$}}                                                           & \textbf{7.6}\textsubscript{\textcolor{darkgreen}{$\downarrow 50\%$}}                                                       \\
                         & Response time & 0.11                                                    & 0.12\textsubscript{\textcolor{darkred}{$\uparrow 9\%$}}                                                                                      & 0.13\textsubscript{\textcolor{darkred}{$\uparrow 18\%$}}                                                  & 0.24\textsubscript{\textcolor{darkred}{$\uparrow 118\%$}}                                                    & \underline{0.11}\textsubscript{\textcolor{darkgreen}{0\%}}                                                          & \textbf{0.05}\textsubscript{\textcolor{darkgreen}{$\downarrow 55\%$}}                                                      \\
                         & Throughput \textcolor{darkgreen}{$\uparrow$}    & 7.4                                                     & \underline{9.2}\textsubscript{\textcolor{darkgreen}{$\uparrow 24\%$}}                                                                                       & 5.5\textsubscript{\textcolor{darkred}{$\downarrow 26\%$}}                                                   & 3.7\textsubscript{\textcolor{darkred}{$\downarrow 50\%$}}                                                     & 7.1\textsubscript{\textcolor{darkred}{$\downarrow 4\%$}}                                                           & \textbf{12.6}\textsubscript{\textcolor{darkgreen}{$\uparrow 70\%$}}                                                      \\ \midrule
\multirow{5}{*}{TextVQA} & Token number  & 576                                                     & 729\textsubscript{\textcolor{darkred}{$\uparrow 27\%$}}                                                                                        & 977\textsubscript{\textcolor{darkred}{$\uparrow 70\%$}}                                                    & 1668\textsubscript{\textcolor{darkred}{$\uparrow 190\%$}}                                                     & \underline{414}\textsubscript{\textcolor{darkgreen}{$\downarrow 28\%$}}                                                           & \textbf{90}\textsubscript{\textcolor{darkgreen}{$\downarrow 84\%$}}                                                        \\
                         & TFLOPs       & 4.76                                                    & \underline{0.31}\textsubscript{\textcolor{darkgreen}{$\downarrow 93\%$}}                                                                                      & 4.14\textsubscript{\textcolor{darkgreen}{$\downarrow 13\%$}}                                                  & 5.09\textsubscript{\textcolor{darkred}{$\uparrow 7\%$}}                                                     & 2.83\textsubscript{\textcolor{darkgreen}{$\downarrow 41\%$}}                                                          & \textbf{0.10}\textsubscript{\textcolor{darkgreen}{$\downarrow 98\%$}}                                                      \\
                         & GPU memory   & 16.0                                                    & 8.4\textsubscript{\textcolor{darkgreen}{$\downarrow 48\%$}}                                                                                       & 40.8\textsubscript{\textcolor{darkred}{$\uparrow 155\%$}}                                                  & 8.8\textsubscript{\textcolor{darkgreen}{$\downarrow 45\%$}}                                                     & \underline{8.1}\textsubscript{\textcolor{darkgreen}{$\downarrow 49\%$}}                                                           & \textbf{7.6}\textsubscript{\textcolor{darkgreen}{$\downarrow 53\%$}}                                                       \\
                         & Response time & 0.11                                                    & 0.12\textsubscript{\textcolor{darkred}{$\uparrow 9\%$}}                                                                                      & 0.56\textsubscript{\textcolor{darkred}{$\uparrow 409\%$}}                                                  & 0.24\textsubscript{\textcolor{darkred}{$\uparrow 118\%$}}                                                    & \underline{0.11}\textsubscript{\textcolor{darkgreen}{$ 0\%$}}                                                          & \textbf{0.05}\textsubscript{\textcolor{darkgreen}{$\downarrow 55\%$}}                                                        \\
                         & Throughput \textcolor{darkgreen}{$\uparrow$}    & 5.4                                                     & \underline{7.1}\textsubscript{\textcolor{darkgreen}{$\uparrow 31\%$}}                                                                                       & 1.4\textsubscript{\textcolor{darkred}{$\downarrow 74\%$}}                                                     & 3.2\textsubscript{\textcolor{darkred}{$\downarrow 41\%$}}                                                       & 5.6\textsubscript{\textcolor{darkgreen}{$\uparrow 4\%$}}                                                           & \textbf{9.8}\textsubscript{\textcolor{darkgreen}{$\uparrow 81\%$}}                                                       \\ \midrule
\multirow{5}{*}{MME}     & Token number  & 576                                                     & 729\textsubscript{\textcolor{darkred}{$\uparrow 27\%$}}                                                                                       & 646\textsubscript{\textcolor{darkred}{$\uparrow 12\%$}}                                                   & 1478\textsubscript{\textcolor{darkred}{$\uparrow 157\%$}}                                                    & \underline{414}\textsubscript{\textcolor{darkgreen}{$\downarrow 28\%$}}                                                             & \textbf{90}\textsubscript{\textcolor{darkgreen}{$\downarrow 84\%$}}                                                          \\
                         & TFLOPs       & 4.43                                                    & \underline{0.30}\textsubscript{\textcolor{darkgreen}{$\downarrow 93\%$}}                                                                                        & 2.72\textsubscript{\textcolor{darkgreen}{$\downarrow 39\%$}}                                                    & 4.45\textsubscript{\textcolor{darkgreen}{$ 0\%$}}                                                      & 2.72\textsubscript{\textcolor{darkgreen}{$\downarrow 39\%$}}                                                            & \textbf{0.09}\textsubscript{\textcolor{darkgreen}{$\downarrow 98\%$}}                                                        \\
                         & GPU memory   & 15.3                                                    & 8.4\textsubscript{\textcolor{darkgreen}{$\downarrow 45\%$}}                                                                                         & 59.2\textsubscript{\textcolor{darkred}{$\uparrow 287\%$}}                                                  & 10.0\textsubscript{\textcolor{darkgreen}{$\downarrow 35\%$}}                                                      & \underline{7.8}\textsubscript{\textcolor{darkgreen}{$\downarrow 49\%$}}                                                             & \textbf{7.6}\textsubscript{\textcolor{darkgreen}{$\downarrow 50\%$}}                                                         \\
                         & Response time & 0.11                                                    & 0.12\textsubscript{\textcolor{darkred}{$\uparrow 9\%$}}                                                                                      & 0.41\textsubscript{\textcolor{darkred}{$\uparrow 273\%$}}                                                  & 0.21\textsubscript{\textcolor{darkred}{$\uparrow 91\%$}}                                                    & \underline{0.11}\textsubscript{\textcolor{darkgreen}{$0\%$}}                                                          & \textbf{0.05}\textsubscript{\textcolor{darkgreen}{$\downarrow 55\%$}}                                                        \\
                         & Throughput \textcolor{darkgreen}{$\uparrow$}    & 5.6                                                     & \underline{9.1}\textsubscript{\textcolor{darkgreen}{$\uparrow 63\%$}}                                                                                       & 1.8\textsubscript{\textcolor{darkred}{$\downarrow 68\%$}}                                                     & 3.7\textsubscript{\textcolor{darkred}{$\downarrow 34\%$}}                                                       & 7.3\textsubscript{\textcolor{darkgreen}{$\uparrow 30\%$}}                                                           & \textbf{11.6}\textsubscript{\textcolor{darkgreen}{$\uparrow 107\%$}}                                                      \\ \midrule
\multirow{5}{*}{MMB}     & Token number  & 576                                                     & 729\textsubscript{\textcolor{darkred}{$\uparrow 27\%$}}                                                                                       & \underline{385}\textsubscript{\textcolor{darkgreen}{$\downarrow 33\%$}}                                                     & 1296\textsubscript{\textcolor{darkred}{$\uparrow 125\%$}}                                                    & 414\textsubscript{\textcolor{darkgreen}{$\downarrow 28\%$}}                                                             & \textbf{90}\textsubscript{\textcolor{darkgreen}{$\downarrow 84\%$}}                                                          \\
                         & TFLOPs       & 4.77                                                    & \underline{0.31}\textsubscript{\textcolor{darkgreen}{$\downarrow 94\%$}}                                                                                        & 0.84\textsubscript{\textcolor{darkgreen}{$\downarrow 82\%$}}                                                    & 3.98\textsubscript{\textcolor{darkgreen}{$\downarrow 17\%$}}                                                      & 1.98\textsubscript{\textcolor{darkgreen}{$\downarrow 58\%$}}                                                            & \textbf{0.09}\textsubscript{\textcolor{darkgreen}{$\downarrow 98\%$}}                                                        \\
                         & GPU memory   & 16.4                                                    & 8.4\textsubscript{\textcolor{darkgreen}{$\downarrow 49\%$}}                                                                                         & \underline{8.0}\textsubscript{\textcolor{darkgreen}{$\downarrow 51\%$}}                                                     & 11.5\textsubscript{\textcolor{darkgreen}{$\downarrow 30\%$}}                                                      & 8.7\textsubscript{\textcolor{darkgreen}{$\downarrow 47\%$}}                                                             & \textbf{7.6}\textsubscript{\textcolor{darkgreen}{$\downarrow 54\%$}}                                                         \\
                         & Response time & 0.11                                                    & 0.12\textsubscript{\textcolor{darkred}{$\uparrow 9\%$}}                                                                                      & \underline{0.09}\textsubscript{\textcolor{darkgreen}{$\downarrow 18\%$}}                                                    & 0.19\textsubscript{\textcolor{darkred}{$\uparrow 73\%$}}                                                    & 0.11\textsubscript{\textcolor{darkgreen}{$0\%$}}                                                          & \textbf{0.05}\textsubscript{\textcolor{darkgreen}{$\downarrow 55\%$}}                                                        \\
                         & Throughput \textcolor{darkgreen}{$\uparrow$}    & 6.1                                                     & \underline{8.4}\textsubscript{\textcolor{darkgreen}{$\uparrow 38\%$}}                                                                                       & 7.0\textsubscript{\textcolor{darkgreen}{$\uparrow 15\%$}}                                                   & 4.7\textsubscript{\textcolor{darkred}{$\downarrow 23\%$}}                                                       & 5.1\textsubscript{\textcolor{darkred}{$\downarrow 16\%$}}                                                             & \textbf{11.7}\textsubscript{\textcolor{darkgreen}{$\uparrow 92\%$}}                                                      \\ \midrule
\multirow{5}{*}{POPE}    & Token number  & 576                                                     & 729\textsubscript{\textcolor{darkred}{$\uparrow 27\%$}}                                                                                       & \underline{353}\textsubscript{\textcolor{darkgreen}{$\downarrow 39\%$}}                                                     & 1666\textsubscript{\textcolor{darkred}{$\uparrow 189\%$}}                                                    & 414\textsubscript{\textcolor{darkgreen}{$\downarrow 28\%$}}                                                             & \textbf{90}\textsubscript{\textcolor{darkgreen}{$\downarrow 84\%$}}                                                          \\
                         & TFLOPs       & 4.40                                                    & \underline{0.30}\textsubscript{\textcolor{darkgreen}{$\downarrow 93\%$}}                                                                                        & 1.51\textsubscript{\textcolor{darkgreen}{$\downarrow 66\%$}}                                                    & 4.99\textsubscript{\textcolor{darkred}{$\uparrow 13\%$}}                                                    & 2.71\textsubscript{\textcolor{darkgreen}{$\downarrow 38\%$}}                                                            & \textbf{0.08}\textsubscript{\textcolor{darkgreen}{$\downarrow 98\%$}}                                                        \\
                         & GPU memory   & 15.3                                                    & 8.4\textsubscript{\textcolor{darkgreen}{$\downarrow 45\%$}}                                                                                         & 7.8\textsubscript{\textcolor{darkgreen}{$\downarrow 49\%$}}                                                     & \underline{7.4}\textsubscript{\textcolor{darkgreen}{$\downarrow 52\%$}}                                                       & 7.8\textsubscript{\textcolor{darkgreen}{$\downarrow 49\%$}}                                                             & \textbf{7.6}\textsubscript{\textcolor{darkgreen}{$\downarrow 50\%$}}                                                         \\
                         & Response time & 0.11                                                    & 0.12\textsubscript{\textcolor{darkred}{$\uparrow 9\%$}}                                                                                      & 0.13\textsubscript{\textcolor{darkred}{$\uparrow 18\%$}}                                                  & 0.23\textsubscript{\textcolor{darkred}{$\uparrow 109\%$}}                                                    & \underline{0.11}\textsubscript{\textcolor{darkgreen}{$ 0\%$}}                                                          & \textbf{0.05}\textsubscript{\textcolor{darkgreen}{$\downarrow 55\%$}}                                                        \\
                         & Throughput \textcolor{darkgreen}{$\uparrow$}    & 7.8                                                     & \underline{9.4}\textsubscript{\textcolor{darkgreen}{$\uparrow 21\%$}}                                                                                       & 5.0\textsubscript{\textcolor{darkred}{$\downarrow 36\%$}}                                                     & 3.8\textsubscript{\textcolor{darkred}{$\downarrow 51\%$}}                                                       & 7.3\textsubscript{\textcolor{darkred}{$\downarrow 6\%$}}                                                             & \textbf{13.2}\textsubscript{\textcolor{darkgreen}{$\uparrow 69\%$}}                                                      \\ \midrule
\multirow{5}{*}{Average} & Token number  & 576                                                     & 729\textsubscript{\textcolor{darkred}{$\uparrow 27\%$}}                                                                                       & 466\textsubscript{\textcolor{darkgreen}{$\downarrow 19\%$}}                                                   & 1561\textsubscript{\textcolor{darkred}{$\uparrow 171\%$}}                                                    & \underline{414}\textsubscript{\textcolor{darkgreen}{$\downarrow 28\%$}}                                                             & \textbf{90}\textsubscript{\textcolor{darkgreen}{$\downarrow 84\%$}}                                                          \\
                         & TFLOPs       & 4.55                                                    & \underline{0.30}\textsubscript{\textcolor{darkgreen}{$\downarrow 93\%$}}                                                                                        & 2.14\textsubscript{\textcolor{darkgreen}{$\downarrow 53\%$}}                                                    & 4.72\textsubscript{\textcolor{darkred}{$\uparrow 4\%$}}                                                    & 2.59\textsubscript{\textcolor{darkgreen}{$\downarrow 43\%$}}                                                            & \textbf{0.09}\textsubscript{\textcolor{darkgreen}{$\downarrow 98\%$}}                                                        \\
                         & GPU memory   & 15.7                                                    & 8.4\textsubscript{\textcolor{darkgreen}{$\downarrow 46\%$}}                                                                                         & 30.8\textsubscript{\textcolor{darkred}{$\uparrow 96\%$}}                                                  & 9.4\textsubscript{\textcolor{darkgreen}{$\downarrow 40\%$}}                                                       & \underline{8.0}\textsubscript{\textcolor{darkgreen}{$\downarrow 49\%$}}                                                             & \textbf{7.6}\textsubscript{\textcolor{darkgreen}{$\downarrow 52\%$}}                                                         \\
                         & Response time & 0.11                                                    & 0.12\textsubscript{\textcolor{darkred}{$\uparrow 9\%$}}                                                                                      & 0.26\textsubscript{\textcolor{darkred}{$\uparrow 136\%$}}                                                  & 0.22\textsubscript{\textcolor{darkred}{$\uparrow 100\%$}}                                                    & \underline{0.11}\textsubscript{\textcolor{darkgreen}{$0\%$}}                                                          & \textbf{0.05}\textsubscript{\textcolor{darkgreen}{$\downarrow 55\%$}}                                                        \\
                         & Throughput \textcolor{darkgreen}{$\uparrow$}    & 6.5                                                     & \underline{8.6}\textsubscript{\textcolor{darkgreen}{$\uparrow 32\%$}}                                                                                       & 4.1\textsubscript{\textcolor{darkred}{$\downarrow 37\%$}}                                                     & 3.8\textsubscript{\textcolor{darkred}{$\downarrow 42\%$}}                                                       & 6.5\textsubscript{\textcolor{darkgreen}{$ 0\%$}}                                                             & \textbf{11.8}\textsubscript{\textcolor{darkgreen}{$\uparrow 82\%$}}                                                      \\ \bottomrule
\end{tabular}}
\label{inference effeciency}
\vspace{-0.3cm} 
\end{table*}

\begin{figure}
    \centering
    \setlength{\belowcaptionskip}{0.3cm}
    \includegraphics[width=1\linewidth]{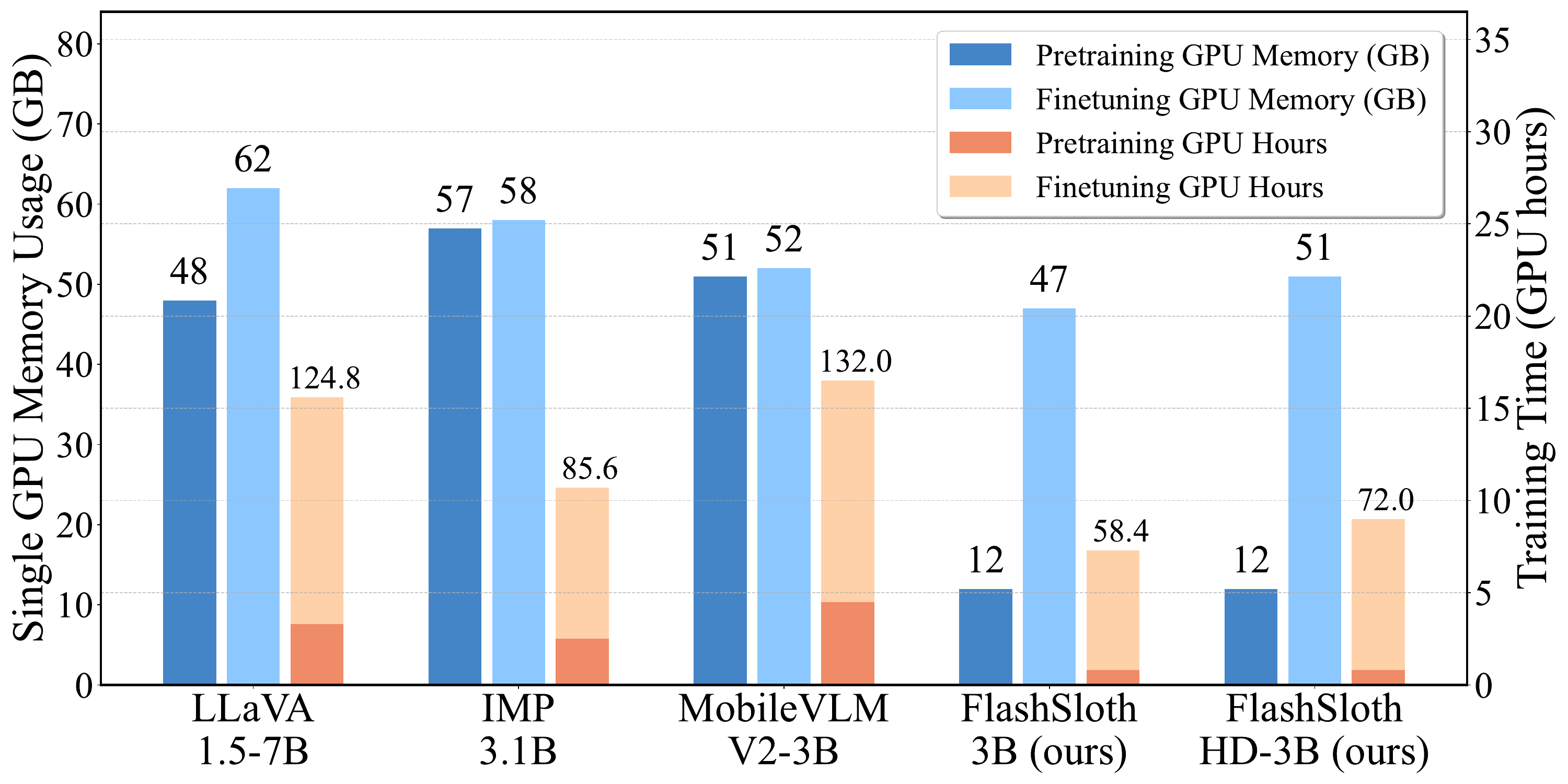}
    \caption{ Comparison between FlashSloth and three MLLMs in terms of training efficiency. The results are obtained using \emph{LLaVA-665k~\cite{llava1.5}} for fair comparisons. FlashSloth is superior in both GPU memory overhead and training time costs.}
    \label{line2}
    \vspace{-0.5cm}
\end{figure}
\subsection{Quantitative Analysis}
\subsubsection{Efficiency Comparison with Existing MLLMs}

We first compare the efficiency of FlashSloth with advanced tiny MLLMs, and also use the representative MLLM LLaVA-1.5-7B~\cite{llava1.5} as reference.\\
\noindent\textbf{Inference Efficiency.} We first compare the inference efficiency of FlashSloth with three advanced tiny MLLMs~\cite{wang2024qwen2,internvl,shao2024imp} in Tab.~\ref{inference effeciency}. For better comparisons, we use LLaVA-1.5-7B~\cite{llava1.5} as reference. From these statics, we can first observe that tiny MLLMs have a lower requirement of GPU memory than LLaVA due to the use of much smaller LLMs. Likewise, their theoretical computation (FLOPS) is also less than LLaVA. However, their actual inferences are not obviously faster, \emph{i.e.}, the response time or throughput. To explain, the large number of visual tokens will greatly slow down the decoding of first token, \emph{i.e.}, response time, making their advantages in \emph{KV caching}~\cite{kvcache} based decoding become not that obvious, especially that the answers of VL examples are often short~\cite{textvqa, chartqa, docvqa}. We can also see that the dynamic resolution designs of Qwen2-VL and InternVL can help to adjust the number of visual tokens for different images, but still keeps a relatively large number, which also result in large latency. Lastly, we can see that with about 80-89\% fewer tokens, FlashSloth exhibits obvious merits than these MLLMs in terms of all inference metrics. For instance, its response time is about 2 and 5 times faster than LLaVA and InternVL, respectively. 
In terms of FlashSloth-HD, its overall efficiency is also superior than the compared MLLMs, even through it uses more visual tokens than FlashSloth. These results well confirm the advantages of FlashSloth in inference and visual compression.

\begin{table*} 
\centering
\renewcommand{\arraystretch}{1.0}
\fontsize{25}{28}\selectfont
\caption{Performance comparison between FlashSloth and the latest tiny MLLMs on seven general multimodal benchmarks and seven common visual-language benchmarks. The best results are \textbf{bold} and the second-best results
are \underline{underlined}.
}
\resizebox{\textwidth}{!}{ 
\begin{tabular}{@{}lcccccccccccccccc@{}}
\toprule[2.5pt]
\multicolumn{1}{l|}{\multirow{2}{*}{\textbf{Model}}}                                     & \multicolumn{1}{c|}{\multirow{2}{*}{\textbf{Params}}} & \multicolumn{1}{c|}{\multirow{2}{*}{\textbf{Data}}} & \multicolumn{7}{c|}{\textbf{Multimodal benchmarks for MLLM}}                                                                 & \multicolumn{7}{c}{\textbf{Common vision-language benchmarks}}                                                \\ \cmidrule(l){4-17} 
\multicolumn{1}{l|}{}                                                           & \multicolumn{1}{c|}{}                        & \multicolumn{1}{c|}{}                               & POPE          & \multicolumn{1}{l}{MME} & MMB           & MM-Vet        & SEED$^{I}$  &MMMU   & \multicolumn{1}{c|}{MathVista}          & GQA           & SQA           & TextVQA       & AI2D          & ChartQA       & DocVQA        & RealWorldQA   \\ \midrule
\rowcolor{gray!20}
\multicolumn{17}{c}{\textbf{MLLMs (LLM \textgreater  3B)}}                                                                                                                                                                                                                                                                                                                                                                           \\ \midrule
\rowcolor{gray!20}
\multicolumn{1}{l|}{LLaVA-1.5~\cite{llava1.5}}            & \multicolumn{1}{c|}{7B}                      & \multicolumn{1}{c|}{1.2M}                           & 85.9          & 1511                    & 64.3          & 30.5          & 58.6     &-     & \multicolumn{1}{c|}{-}             & 62.0          & 66.8          & 58.2          & -             & -             & -             & -             \\
\rowcolor{gray!20}
\multicolumn{1}{l|}{LLaVA-NeXT~\cite{llavanext}}          & \multicolumn{1}{c|}{8B}                      & \multicolumn{1}{c|}{1.4M}                               & 86.5          & 1519                    & 67.4          & -             & 70.2     &35.8     & \multicolumn{1}{c|}{34.6}          & 64.2          & 70.1          & 64.9          & -             & -             & -             & -             \\
\rowcolor{gray!20}
\multicolumn{1}{l|}{Eagle-X5~\cite{eagle}}                & \multicolumn{1}{c|}{7B}                      & \multicolumn{1}{c|}{1.2M}                           & 88.8          & 1528                    & 68.4          & -             & 73.9     &36.3     & \multicolumn{1}{c|}{37.0}          & 64.9          & 70.0          & 71.2          & -             & 67.7          & -             & -             \\
\rowcolor{gray!20}
\multicolumn{1}{l|}{DeepSeek-VL~\cite{lu2024deepseek}}                                                & \multicolumn{1}{c|}{7B}                     & \multicolumn{1}{c|}{\textgreater{}100M}                               & 88.1          & -                       & 73.2          & 41.5          & 70.4      &36.6    & \multicolumn{1}{c|}{-}          & -             & -             & -             & -             & -             & -             & -             \\ 

\midrule
\multicolumn{17}{c}{\textbf{Lightweight MLLMs (LLM \textless  3B)}}                                                                                                                                                                                                                                                                                                                                                                  \\ \midrule
\multicolumn{1}{l|}{MobileVLM-V2~\cite{chu2024mobilevlm}} & \multicolumn{1}{c|}{1.7B}                    & \multicolumn{1}{c|}{3.6M}                           & 84.3          & 1302                    & 57.7          & -             & -       &-      & \multicolumn{1}{c|}{-}             & 59.3          & 66.7          & 52.1          & -             & -             & -             & -             \\
\multicolumn{1}{l|}{LLaVA-Phi~\cite{llavaphi}}            & \multicolumn{1}{c|}{3B}                      & \multicolumn{1}{c|}{1.2M}                           & 85.0          & 1335                    & 59.8          & -             & -        &-     & \multicolumn{1}{c|}{-}             & -             & 68.4          & 48.6          & -             & -             & -             & -             \\
\multicolumn{1}{l|}{MobileVLM-V2~\cite{chu2024mobilevlm}} & \multicolumn{1}{c|}{3B}                      & \multicolumn{1}{c|}{3.6M}                           & 84.7          & 1441                    & 63.2          & -             & -       &-      & \multicolumn{1}{c|}{-}             & 61.1          & 70.0          & 57.5          & -             & -             & -             & -             \\
\multicolumn{1}{l|}{DeepSeek-VL~\cite{lu2024deepseek}}                  & \multicolumn{1}{c|}{1.7B}                    & \multicolumn{1}{c|}{\textgreater{}100M}             & 87.6          & 1532                    & 64.6          & 34.8          & 66.7     &32.2     & \multicolumn{1}{c|}{31.1}          & -             & -             & -             & -             & -             & -             & 49.7          \\
\multicolumn{1}{l|}{mini-gemini~\cite{minigemini}}        & \multicolumn{1}{c|}{2.2B}                    & \multicolumn{1}{c|}{2.7M}                           & -             & 1653                    & 59.8          & 31.1          & -        &31.7     & \multicolumn{1}{c|}{29.4}          & -             & -             & 56.2          & -             & -             & 34.2          & -             \\
\multicolumn{1}{l|}{PaliGemma~\cite{team2024gemma}}             & \multicolumn{1}{c|}{3B}                      & \multicolumn{1}{c|}{\textgreater 1B}                & -             & 1686                    & 71.0          & 33.1          & 69.6       &34.9   & \multicolumn{1}{c|}{28.7}          & -             & -             & 68.1          & 68.3          & -             & 34.2          & -             \\
\multicolumn{1}{l|}{MiniCPM-V-2~\cite{minicpm}}           & \multicolumn{1}{c|}{2.8B}                    & \multicolumn{1}{c|}{\textgreater 500M}              & 87.8          & 1809                    & 69.6          & 41.0          & 67.1      &38.2   & \multicolumn{1}{c|}{38.7}          & -             & 80.7          & 74.1          & 62.9          & 59.8          & 71.9          & 55.8          \\
\multicolumn{1}{l|}{Imp-v1~\cite{shao2024imp}}            & \multicolumn{1}{c|}{3B}                      & \multicolumn{1}{c|}{1.2M}                           & \underline {88.0}    & 1434                    & 66.5          & -             & -       &-      & \multicolumn{1}{c|}{-}             & -             & 70.0          & 59.4          & -             & -             & -             & -             \\

\multicolumn{1}{l|}{InternVL1.5~\cite{internvl}}        & \multicolumn{1}{c|}{2.2B}                    & \multicolumn{1}{c|}{\textgreater 5M}                &               & \textbf{1902}           & 70.9          & 39.3          & 69.8     &34.6     & \multicolumn{1}{c|}{41.1}          & \underline {61.6}    & 84.9          & 70.5          & 69.8          & \underline {74.8}    & 85.0          & -             \\
\multicolumn{1}{l|}{InternVL2~\cite{internvl}}           & \multicolumn{1}{c|}{2.2B}                    & \multicolumn{1}{c|}{\textgreater 5M}                & 85.2          & \underline {1877}              & 73.2          & 44.6          &\underline {71.6}  &36.3  & \multicolumn{1}{c|}{\textbf{46.3}}          & -             & \textbf{94.1} & 73.4          & 74.1          & \textbf{76.2} & 86.9          & 57.3          \\

\multicolumn{1}{l|}{MM1.5~\cite{mm1}}                     & \multicolumn{1}{c|}{3B}                      & \multicolumn{1}{c|}{\textgreater 45M}               & \textbf{88.1} & 1798                    & -             & 41.0          & \textbf{72.4} &37.1 & \multicolumn{1}{c|}{\underline{44.4}}          & -             & 82.1          &\underline {76.5}    & 65.7          & 74.2          & \underline {87.7}    & 56.9          \\ 
\multicolumn{1}{l|}{Qwen2-VL~\cite{wang2024qwen2}}              & \multicolumn{1}{c|}{2B}                      & \multicolumn{1}{c|}{-}                              & -             & 1872                    & \underline {74.9}    & \textbf{49.5} & -       &\textbf{41.1}      & \multicolumn{1}{c|}{43.0} & -             & -             & \textbf{79.7} & \underline {74.7}    & 73.5          & \textbf{90.1} & \textbf{62.9} \\

\midrule
\multicolumn{1}{l|}{FlashSloth (ours)}                                          & \multicolumn{1}{c|}{3.2B}                    & \multicolumn{1}{c|}{3.7M}                           & 86.3          & 1702                    & 73.0          & 41.9          & 68.0      &\underline{39.7}    & \multicolumn{1}{c|}{42.5}    & 61.1          & 88.6          & 64.6          & 72.5          & 51.0          & 48.6          & 54.8          \\
\multicolumn{1}{l|}{FlashSloth-HD (ours)}                                       & \multicolumn{1}{c|}{3.2B}                    & \multicolumn{1}{c|}{3.7M}                           & 87.2          & 1745                    & \textbf{75.7} & \underline{49.0}    &71.2        &37.8       & \multicolumn{1}{c|}{40.6}          & \textbf{62.5} & \underline {91.1}    & 71.0          & \textbf{75.3} & 69.8          & 74.8          & \underline {59.9}    \\ \bottomrule[2.5pt]
\end{tabular}
}

\label{performance}
\vspace{-0.3cm}
\end{table*}

\noindent\textbf{Training Efficiency.} We further report the training expenditures of FlashSloth and the other four MLLMs in Fig.~\ref{line2}. For a quick comparison. we use the pretraining and tuning splits of LLaVA~\cite{llava1.5}, and the per-GPU batch size is set to 32 for pretraining and 8 for instruction tuning. From these plots, we can first find that FlashSloth consumes much less training time than the other MLLMs, especially pretraining. In practice, its pretraining using the LLaVA split only takes 6.4 GPU hours, about 76\% and 68\% less than LLaVA and IMP, respectively. Its SFT tuning time (52 GPU hours) is longer due to more examples used, and it is also slightly affected by the input queries. However, the cost is still lower than than IMP (65.6 GPU hours) and MobileVLM (96 GPU hours). Similarly, with a well designed training scheme, FlashSloth-HD has a slightly longer training time (6.4+65.6 GPU hours), which is still cheaper than the other MLLMs. The other observation from Fig.~\ref{line2} is that tiny MLLMs require GPU memories close to that of LLaVA except our FlashSloth. During pretraining, both FlashSloth and FlashSloth-HD only use about 81 visual tokens, making its memory overhead much lower than the other MLLMs. During instruction tuning, their GPU memories increase greatly. To explain, the queries are given for each instruction, and an SFT example is often a multi-round conversation, so the multiple paddings will bring in a larger sequence. With effective compression, this overhead is still lower than the compared methods. Overall, these results show the merits of FlashSloth in training efficiency.

\subsubsection{Performance Comparison with Existing MLLMs}
We make performance comparisons between FlashSloth-HD and a bunch of advanced tiny MLLMs on 14 highly-competitive VL and MLLM benchmarks, as shown in Table~\ref{performance}. From this table, we can first observe that FlashSloth is very competitive on common VL tasks with a much smaller number of visual tokens. For instance, its performance on MMB~\cite{mmb}, GQA~\cite{gqa} and SQA~\cite{sqa} is much better than several previous tiny MLLMs with similar training data amount, \emph{e.g.}, MobileVLM~\cite{chu2024mobilevlm}, Mini-Gemini~\cite{minigemini} and Imp~\cite{shao2024imp}. Compared to the SOTA tiny MLLMs, such as InternVL~\cite{internvl} and Qwen2-VL~\cite{wang2024qwen2}, FlashSloth also exhibits good competitiveness on these tasks, but it still lags behind them on the OCR tasks like DocVQA~\cite{docvqa}, which requires high-resolution image inputs. In this case, we can see that its HD version, \emph{i.e.}, FlashSloth-HD, can well compensate this shortcoming. Overall, retaining better efficiency, FlashSloth-HD can generally reach the capability of InternVL2, and well shorten its gap to Qwen2-VL. Moreover, FlashSloth can even achieve new SOTA performance among tiny MLLMs on several benchmarks, such as MMB, GQA and AI2D. Considering the much smaller amount of training data for FlashSloth-HD, these results are in fact very notable, well confirming our motivation and designs.

\begin{table}
\centering
\caption{Ablation study on the component design of FlashSloth, initialization method of query tokens, and the number of query tokens, conducted on four benchmarks and TFLOPs. The method marked with \ddag~ represents our final selected setting.
}
\setlength{\belowcaptionskip}{0.cm}
\resizebox{1\columnwidth}{!}{
\setlength{\tabcolsep}{1pt}
\renewcommand{\arraystretch}{0.8} 
\setlength{\arrayrulewidth}{0.2pt}
{\fontsize{4.5}{5}\selectfont
\vspace{-0.1cm}
\begin{tabular}{@{}cccccc@{}}
\toprule[0.4pt]
\multicolumn{1}{c|}{\textbf{Choices}}   & \textbf{GQA} & \textbf{POPE} & \textbf{MME} & \multicolumn{1}{c|}{\textbf{MMB}} & \textbf{TFLOPs} \\ \midrule[0.2pt]
\multicolumn{6}{c}{Visual compression designs}                                                                                                  \\ \midrule[0.2pt]
\multicolumn{1}{c|}{baseline (729)}           & 61.8         & 87.8          & 1491.4       & \multicolumn{1}{c|}{67.9}         &0.30                 \\
\multicolumn{1}{c|}{+ Avg.Pool (81)}         & 59.6         & 84.4          & 1440.3       & \multicolumn{1}{c|}{62.7}         &0.05                 \\
\multicolumn{1}{c|}{+ Att.Pool (81)}         & 60.2         & 86.4          & 1444.8       & \multicolumn{1}{c|}{63.9}         &0.06                 \\
\multicolumn{1}{c|}{+ Att.Pool \& EmbQ\ddag (90)} & 60.8         & 87.8          & 1491.0       & \multicolumn{1}{c|}{67.9}         &0.09                 \\ \midrule[0.2pt]
\multicolumn{6}{c}{Query initialization.}                                                                                                    \\ \midrule[0.2pt]
\multicolumn{1}{c|}{Fixed Dot Token}    & 60.6         & 86.4          & 1423.0       & \multicolumn{1}{c|}{66.1}         &0.09                 \\
\multicolumn{1}{c|}{Random Init}        & 60.5         & 86.9          & 1469.7       & \multicolumn{1}{c|}{65.6}         &0.09                 \\
\multicolumn{1}{c|}{Dot init\ddag}           & 60.8         & 87.8          & 1491.0       & \multicolumn{1}{c|}{67.9}         &0.09                 \\ \midrule[0.2pt]
\multicolumn{6}{c}{Number of queries in Emb.Q}                                                                                                  \\ \midrule[0.2pt]
\multicolumn{1}{c|}{0}                  & 60.3         & 86.4          & 1444.8       & \multicolumn{1}{c|}{63.9}         &0.06                 \\
\multicolumn{1}{c|}{6}                  & 60.5         & 86.8          & 1453.7       & \multicolumn{1}{c|}{65.8}         &0.08                 \\
\multicolumn{1}{c|}{9\ddag}                  & 60.8         & 87.8          & 1491.0       & \multicolumn{1}{c|}{67.9}         & 0.09                \\
\multicolumn{1}{c|}{12}                 & 60.7         & 86.9          & 1437.3       & \multicolumn{1}{c|}{67.8}         &0.10                 \\
\multicolumn{1}{c|}{15}                 & 60.8         & 87.2          & 1468.5       & \multicolumn{1}{c|}{67.0}         &0.11                 \\ \bottomrule[0.4pt]
\end{tabular}}
}
\label{ablation1}
\vspace{-0.2mm}
\end{table}

\begin{figure*}[ht]
    \centering
    \begin{subfigure}[b]{\textwidth} 
        \centering
        \includegraphics[width=\textwidth]{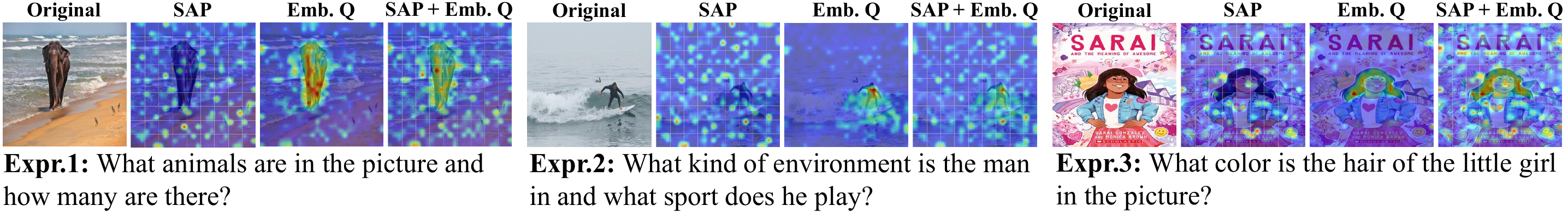}
        \caption{\textbf{Visualization of attention maps for SAP and Emb.Q in FlashSloth.} }
        \label{fig:subfig1}
        \vspace{+1.5mm}
    \end{subfigure}
    \begin{subfigure}[b]{\textwidth}  
        \centering
        \includegraphics[width=\textwidth]{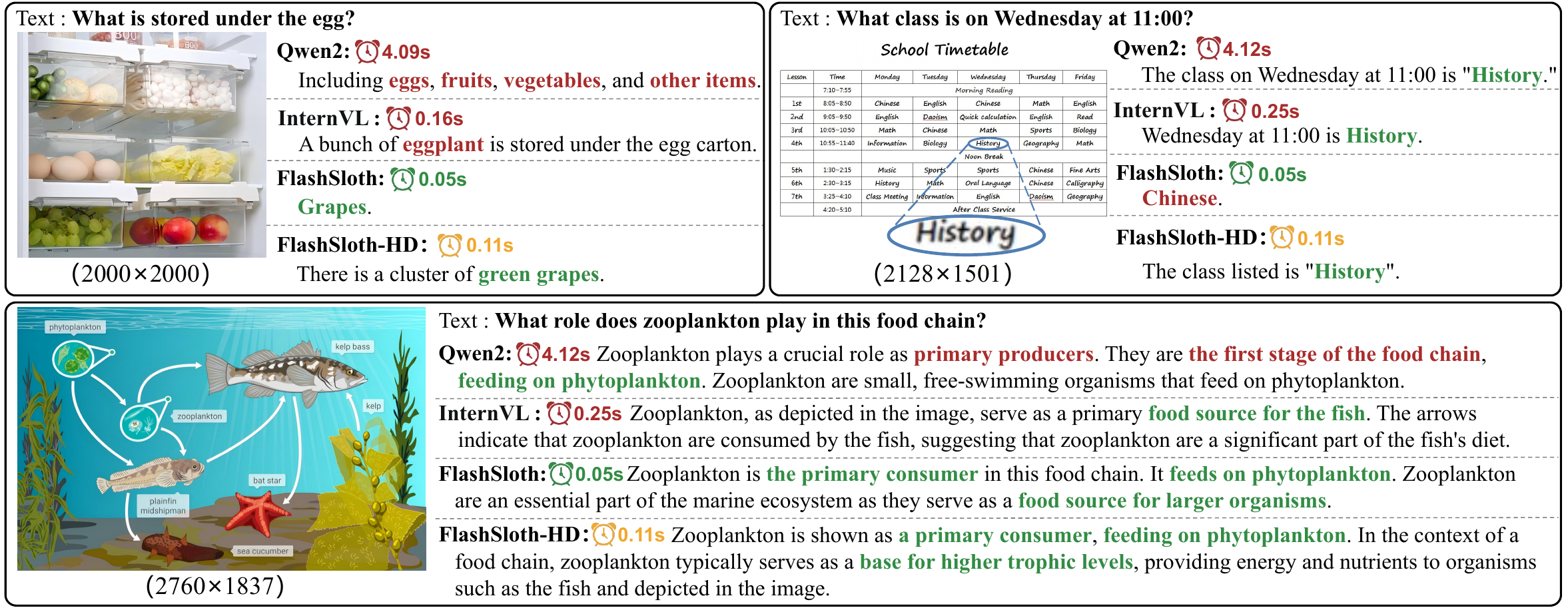}
        \vspace{-1.2mm}
        \setlength{\abovecaptionskip}{0.cm}
        \caption{\textbf{Predictions of \textbf{FlashSloth}, FlashSloth-HD and SOTA  tiny MLLMs.}}
        \label{fig:subfig2}
    \end{subfigure}
    \setlength{\abovecaptionskip}{0.cm}
    \caption{Visualized results of FlashSloth with Qwen2-VL-2B and InternVL2-2.2B. Subfigure-(a) show the attention maps of different visual compressions for FlashSloth, which shows  the abilities of SAP in visual saliency compression and EmbQ  for instruction-related visual querying. Subfigure-(b) shows FlashSloth's rapid response time and its performance on common tasks, which is comparable to or better than the SOTA tiny MLLMs. The clock time represents the response of the model. Incorrect answers are in \textcolor[HTML]{B02425}{RED}.}
    \setlength{\belowcaptionskip}{0.cm}
    \vspace{-0.1cm}
    \label{fig:main_fig}
\end{figure*}

\subsubsection{Ablation Study}
\vspace{-0.2cm}
In the first block of Tab. \ref{ablation1}, we first compare different visual compression designs. The most simple solution is \emph{average pooling}, but it tends to lose key visual information, leading to obvious declines in most benchmarks, \emph{e.g.}, MME and MMB for multimodal perception and recognition. In contrast, our SAP can well keep the visual saliency, so as to obtaining better performance than simple pooling. In addition, we can also see that with the combination of EmbQ and SAP, FlashSloth can obtain very marginal performance drops compared to the baseline, while its efficiency is much better. This result confirms the supplement of EmbQ to SAP.
In the second and last blocks of Tab. \ref{ablation1}, we examine the settings of EmbQ. We can first see that the initialization of queries has impact on EmbQ. The random initialization can serves the target of EmbQ, but the initialization of text \emph{dot} token further improves performance slightly, suggesting their better interactions with other input tokens in MLLMs. Besides, we can also see that the number of queries required by EmbQ is very small, and 9 tokens are enough for visual querying. To explain, EmbQ serves to capture instruction-related information at a coarse granularity, as discussed above. As a supplement to SAP, this design only requires a few queries, especially considering the short questions in MLLM tasks. Overall, these results well confirm the designs of EmbQ.

\subsection{Qualitative Analysis}

To gain deeper insight into FlashSloth's process of enhancing visual feature description, we visualize the attention results of SAP and EmbQ, as shown in the figure~\ref{fig:subfig1}. As observed, SAP distributes attention more broadly, allowing the model to focus on salient information from different regions of the image. This helps the model capture salient details in images , such as the three small birds in the left image, seabirds in the middle image, and tiny text in the right image. In contrast, the embedded query focuses more narrowly on key, text-related information in the image, such as the elephant in the left image, the surfer in the middle image, and the hair in the right image. By combining these two attention mechanisms, the model can effectively prioritize the most important information in the image, enhancing the expressiveness of visual features. This demonstrates that the synergy between SAP and EmbQ allows FlashSloth to fully leverage visual information, resulting in improved performance across multi-task scenarios.

In the Figure~\ref{fig:subfig2}, We visualize the predictions  of FlashSloth , FlashSloth-HD, Qwen2-VL and InternVL-2 for different VL examples. First, FlashSloth exhibits extremely fast response times, with latency significantly lower than the other two models, providing a good user experience. Second, for coarse-grained real-world QA and scientific QA tasks, FlashSloth's performance is on par with or even surpasses that of the other two models. In the top-left example, FlashSloth identifies grapes that the other models miss, and FlashSloth-HD answers in more detail. In the bottom example, FlashSloth provides the most accurate answer to a biological question. However, due to its lower resolution, FlashSloth underperforms on the OCR tasks. For instance, in the top-right example, FlashSloth fails to recognize correctly, but upon increasing the input resolution, FlashSloth-HD handles fine-grained OCR tasks effectively.
\section{Conclusion}
In this paper, we introduce FlashSloth, a powerful and fast tiny MLLM. By incorporating effective embedded visual compression designs, FlashSloth effectively captures both visual saliency and instruction-related semantics, achieving an optimal balance between performance and efficiency. Extensive comparisons with existing tiny MLLMs on various benchmarks demonstrate that FlashSloth significantly enhances both training and inference efficiency while maintaining competitive performance, which well validates its motivation and designs. 
\section{Acknowledgments}
This work was supported by National Science and Technology Major Project (No. 2022ZD0118201), the National Science Fund for Distinguished Young Scholars (No. 62025603), the National Natural Science Foundation of China (No. U21B2037, No. U22B2051, No. 623B2088, No. U23A20383, No. U21A20472, No. 62176222, No. 62176223, No. 62176226, No. 62072386, No. 62072387, No. 62072389, No. 62002305 and No. 62272401), and the Natural Science Foundation of Fujian Province of China (No. 2021J06003, No. 2022J06001).
{
    \small
    \bibliographystyle{ieeenat_fullname}
    \bibliography{main}

\begin{thebibliography}{75}
\providecommand{\natexlab}[1]{#1}
\providecommand{\url}[1]{\texttt{#1}}
\expandafter\ifx\csname urlstyle\endcsname\relax
  \providecommand{\doi}[1]{doi: #1}\else
  \providecommand{\doi}{doi: \begingroup \urlstyle{rm}\Url}\fi

\bibitem[Achiam et~al.(2023)Achiam, Adler, Agarwal, Ahmad, Akkaya, Aleman, Almeida, Altenschmidt, Altman, Anadkat, et~al.]{gpt4}
Josh Achiam, Steven Adler, Sandhini Agarwal, Lama Ahmad, Ilge Akkaya, Florencia~Leoni Aleman, Diogo Almeida, Janko Altenschmidt, Sam Altman, Shyamal Anadkat, et~al.
\newblock Gpt-4 technical report.
\newblock \emph{arXiv preprint arXiv:2303.08774}, 2023.

\bibitem[Anderson et~al.(2018)Anderson, He, Buehler, Teney, Johnson, Gould, and Zhang]{bottom}
Peter Anderson, Xiaodong He, Chris Buehler, Damien Teney, Mark Johnson, Stephen Gould, and Lei Zhang.
\newblock Bottom-up and top-down attention for image captioning and visual question answering.
\newblock In \emph{Proceedings of the IEEE conference on computer vision and pattern recognition}, pages 6077--6086, 2018.

\bibitem[Bai et~al.(2023)Bai, Bai, Chu, Cui, Dang, Deng, Fan, Ge, Han, Huang, et~al.]{bai2023Qwen}
Jinze Bai, Shuai Bai, Yunfei Chu, Zeyu Cui, Kai Dang, Xiaodong Deng, Yang Fan, Wenbin Ge, Yu Han, Fei Huang, et~al.
\newblock Qwen technical report.
\newblock \emph{arXiv preprint arXiv:2309.16609}, 2023.

\bibitem[Banks and Warkentin(2024)]{gemma}
Jeanine Banks and Tris Warkentin.
\newblock Gemma: Introducing new state-of-the-art open models.
\newblock \emph{Google. Available online at: https://blog. google/technology/developers/gemma-open-models/(accessed 6 April, 2024)}, 2024.

\bibitem[Brown et~al.(2020)Brown, Mann, Ryder, Subbiah, Kaplan, Dhariwal, Neelakantan, Shyam, Sastry, Askell, Agarwal, Herbert-Voss, Krueger, Henighan, Child, Ramesh, Ziegler, Wu, Winter, Hesse, Chen, Sigler, Litwin, Gray, Chess, Clark, Berner, McCandlish, Radford, Sutskever, and Amodei]{kvcache}
Tom Brown, Benjamin Mann, Nick Ryder, Melanie Subbiah, Jared~D Kaplan, Prafulla Dhariwal, Arvind Neelakantan, Pranav Shyam, Girish Sastry, Amanda Askell, Sandhini Agarwal, Ariel Herbert-Voss, Gretchen Krueger, Tom Henighan, Rewon Child, Aditya Ramesh, Daniel Ziegler, Jeffrey Wu, Clemens Winter, Chris Hesse, Mark Chen, Eric Sigler, Mateusz Litwin, Scott Gray, Benjamin Chess, Jack Clark, Christopher Berner, Sam McCandlish, Alec Radford, Ilya Sutskever, and Dario Amodei.
\newblock Language models are few-shot learners.
\newblock In \emph{Advances in Neural Information Processing Systems}, pages 1877--1901. Curran Associates, Inc., 2020.

\bibitem[Cha et~al.(2024)Cha, Kang, Mun, and Roh]{honeybee}
Junbum Cha, Wooyoung Kang, Jonghwan Mun, and Byungseok Roh.
\newblock Honeybee: Locality-enhanced projector for multimodal llm.
\newblock In \emph{Proceedings of the IEEE/CVF Conference on Computer Vision and Pattern Recognition}, pages 13817--13827, 2024.

\bibitem[Chen et~al.(2024{\natexlab{a}})Chen, Zhao, Liu, Bai, Lin, Zhou, and Chang]{fastv}
Liang Chen, Haozhe Zhao, Tianyu Liu, Shuai Bai, Junyang Lin, Chang Zhou, and Baobao Chang.
\newblock An image is worth 1/2 tokens after layer 2: Plug-and-play inference acceleration for large vision-language models.
\newblock \emph{arXiv preprint arXiv:2403.06764}, 2024{\natexlab{a}}.

\bibitem[Chen et~al.(2015)Chen, Fang, Lin, Vedantam, Gupta, Dollar, and Zitnick]{caption}
Xinlei Chen, Hao Fang, Tsung-Yi Lin, Ramakrishna Vedantam, Saurabh Gupta, Piotr Dollar, and C.~Lawrence Zitnick.
\newblock Microsoft coco captions: Data collection and evaluation server, 2015.

\bibitem[Chen et~al.(2022)Chen, Wang, Changpinyo, Piergiovanni, Padlewski, Salz, Goodman, Grycner, Mustafa, Beyer, et~al.]{chen2022pali}
Xi Chen, Xiao Wang, Soravit Changpinyo, AJ Piergiovanni, Piotr Padlewski, Daniel Salz, Sebastian Goodman, Adam Grycner, Basil Mustafa, Lucas Beyer, et~al.
\newblock Pali: A jointly-scaled multilingual language-image model.
\newblock \emph{arXiv preprint arXiv:2209.06794}, 2022.

\bibitem[Chen et~al.(2024{\natexlab{b}})Chen, Wang, Tian, Ye, Gao, Cui, Tong, Hu, Luo, Ma, Ma, Wang, Dong, Yan, Guo, He, Shi, Jin, Xu, Wang, Wei, Li, Zhang, Zhang, Cai, Wen, Yan, Dou, Lu, Zhu, Lu, Lin, Qiao, Dai, and Wang]{internvl}
Zhe Chen, Weiyun Wang, Hao Tian, Shenglong Ye, Zhangwei Gao, Erfei Cui, Wenwen Tong, Kongzhi Hu, Jiapeng Luo, Zheng Ma, Ji Ma, Jiaqi Wang, Xiaoyi Dong, Hang Yan, Hewei Guo, Conghui He, Botian Shi, Zhenjiang Jin, Chao Xu, Bin Wang, Xingjian Wei, Wei Li, Wenjian Zhang, Bo Zhang, Pinlong Cai, Licheng Wen, Xiangchao Yan, Min Dou, Lewei Lu, Xizhou Zhu, Tong Lu, Dahua Lin, Yu Qiao, Jifeng Dai, and Wenhai Wang.
\newblock How far are we to gpt-4v? closing the gap to commercial multimodal models with open-source suites, 2024{\natexlab{b}}.

\bibitem[Chu et~al.(2023)Chu, Qiao, Lin, Xu, Yang, Hu, Wei, Zhang, Zhang, Wei, et~al.]{chu2023mobilevlm}
Xiangxiang Chu, Limeng Qiao, Xinyang Lin, Shuang Xu, Yang Yang, Yiming Hu, Fei Wei, Xinyu Zhang, Bo Zhang, Xiaolin Wei, et~al.
\newblock Mobilevlm: A fast, strong and open vision language assistant for mobile devices.
\newblock \emph{arXiv preprint arXiv:2312.16886}, 2023.

\bibitem[Chu et~al.(2024)Chu, Qiao, Zhang, Xu, Wei, Yang, Sun, Hu, Lin, Zhang, et~al.]{chu2024mobilevlm}
Xiangxiang Chu, Limeng Qiao, Xinyu Zhang, Shuang Xu, Fei Wei, Yang Yang, Xiaofei Sun, Yiming Hu, Xinyang Lin, Bo Zhang, et~al.
\newblock Mobilevlm v2: Faster and stronger baseline for vision language model.
\newblock \emph{arXiv preprint arXiv:2402.03766}, 2024.

\bibitem[Dai et~al.(2023)Dai, Li, Li, Tiong, Zhao, Wang, Li, Fung, and Hoi]{instructblip}
Wenliang Dai, Junnan Li, Dongxu Li, Anthony Meng~Huat Tiong, Junqi Zhao, Weisheng Wang, Boyang Li, Pascale Fung, and Steven Hoi.
\newblock Instructblip: Towards general-purpose vision-language models with instruction tuning, 2023.

\bibitem[Fu et~al.(2024)Fu, Chen, Shen, Qin, Zhang, Lin, Yang, Zheng, Li, Sun, Wu, and Ji]{mme}
Chaoyou Fu, Peixian Chen, Yunhang Shen, Yulei Qin, Mengdan Zhang, Xu Lin, Jinrui Yang, Xiawu Zheng, Ke Li, Xing Sun, Yunsheng Wu, and Rongrong Ji.
\newblock Mme: A comprehensive evaluation benchmark for multimodal large language models, 2024.

\bibitem[Goyal et~al.(2017)Goyal, Khot, Summers-Stay, Batra, and Parikh]{making}
Yash Goyal, Tejas Khot, Douglas Summers-Stay, Dhruv Batra, and Devi Parikh.
\newblock Making the v in vqa matter: Elevating the role of image understanding in visual question answering.
\newblock In \emph{Proceedings of the IEEE conference on computer vision and pattern recognition}, pages 6904--6913, 2017.

\bibitem[Hu et~al.(2024{\natexlab{a}})Hu, Tu, Han, He, Cui, Long, Zheng, Fang, Huang, Zhao, Zhang, Thai, Zhang, Wang, Yao, Zhao, Zhou, Cai, Zhai, Ding, Jia, Zeng, Li, Liu, and Sun]{hu2024minicpmunveilingpotentialsmall}
Shengding Hu, Yuge Tu, Xu Han, Chaoqun He, Ganqu Cui, Xiang Long, Zhi Zheng, Yewei Fang, Yuxiang Huang, Weilin Zhao, Xinrong Zhang, Zheng~Leng Thai, Kaihuo Zhang, Chongyi Wang, Yuan Yao, Chenyang Zhao, Jie Zhou, Jie Cai, Zhongwu Zhai, Ning Ding, Chao Jia, Guoyang Zeng, Dahai Li, Zhiyuan Liu, and Maosong Sun.
\newblock Minicpm: Unveiling the potential of small language models with scalable training strategies, 2024{\natexlab{a}}.

\bibitem[Hu et~al.(2024{\natexlab{b}})Hu, Tu, Han, He, Cui, Long, Zheng, Fang, Huang, Zhao, et~al.]{minicpm}
Shengding Hu, Yuge Tu, Xu Han, Chaoqun He, Ganqu Cui, Xiang Long, Zhi Zheng, Yewei Fang, Yuxiang Huang, Weilin Zhao, et~al.
\newblock Minicpm: Unveiling the potential of small language models with scalable training strategies.
\newblock \emph{arXiv preprint arXiv:2404.06395}, 2024{\natexlab{b}}.

\bibitem[Hu et~al.(2024{\natexlab{c}})Hu, Li, van~de Weijer, Gao, Khan, Yang, Cheng, Wang, and Wang]{tome}
Taihang Hu, Linxuan Li, Joost van~de Weijer, Hongcheng Gao, Fahad~Shahbaz Khan, Jian Yang, Ming-Ming Cheng, Kai Wang, and Yaxing Wang.
\newblock Token merging for training-free semantic binding in text-to-image synthesis, 2024{\natexlab{c}}.

\bibitem[Hudson and Manning(2019)]{gqa}
Drew~A Hudson and Christopher~D Manning.
\newblock Gqa: A new dataset for real-world visual reasoning and compositional question answering.
\newblock In \emph{Proceedings of the IEEE/CVF conference on computer vision and pattern recognition}, pages 6700--6709, 2019.

\bibitem[Javaheripi et~al.(2023)Javaheripi, Bubeck, Abdin, Aneja, Bubeck, Mendes, Chen, Del~Giorno, Eldan, Gopi, et~al.]{phi}
Mojan Javaheripi, S{\'e}bastien Bubeck, Marah Abdin, Jyoti Aneja, Sebastien Bubeck, Caio C{\'e}sar~Teodoro Mendes, Weizhu Chen, Allie Del~Giorno, Ronen Eldan, Sivakanth Gopi, et~al.
\newblock Phi-2: The surprising power of small language models.
\newblock \emph{Microsoft Research Blog}, 2023.

\bibitem[Jiang et~al.(2020)Jiang, Misra, Rohrbach, Learned-Miller, and Chen]{in}
Huaizu Jiang, Ishan Misra, Marcus Rohrbach, Erik Learned-Miller, and Xinlei Chen.
\newblock In defense of grid features for visual question answering.
\newblock In \emph{Proceedings of the IEEE/CVF conference on computer vision and pattern recognition}, pages 10267--10276, 2020.

\bibitem[Kant et~al.(2021)Kant, Moudgil, Batra, Parikh, and Agrawal]{contrast}
Yash Kant, Abhinav Moudgil, Dhruv Batra, Devi Parikh, and Harsh Agrawal.
\newblock Contrast and classify: Training robust vqa models.
\newblock In \emph{Proceedings of the IEEE/CVF International Conference on Computer Vision}, pages 1604--1613, 2021.

\bibitem[Kazemzadeh et~al.(2014)Kazemzadeh, Ordonez, Matten, and Berg]{rec}
Sahar Kazemzadeh, Vicente Ordonez, Mark Matten, and Tamara Berg.
\newblock Referitgame: Referring to objects in photographs of natural scenes.
\newblock In \emph{Proceedings of the 2014 conference on empirical methods in natural language processing (EMNLP)}, pages 787--798, 2014.

\bibitem[Kembhavi et~al.(2016)Kembhavi, Salvato, Kolve, Seo, Hajishirzi, and Farhadi]{ai2d}
Aniruddha Kembhavi, Mike Salvato, Eric Kolve, Minjoon Seo, Hannaneh Hajishirzi, and Ali Farhadi.
\newblock A diagram is worth a dozen images.
\newblock In \emph{Computer Vision--ECCV 2016: 14th European Conference, Amsterdam, The Netherlands, October 11--14, 2016, Proceedings, Part IV 14}, pages 235--251. Springer, 2016.

\bibitem[Li et~al.(2024{\natexlab{a}})Li, Ge, Ge, Wang, Wang, Zhang, and Shan]{seed}
Bohao Li, Yuying Ge, Yixiao Ge, Guangzhi Wang, Rui Wang, Ruimao Zhang, and Ying Shan.
\newblock Seed-bench: Benchmarking multimodal large language models.
\newblock In \emph{Proceedings of the IEEE/CVF Conference on Computer Vision and Pattern Recognition}, pages 13299--13308, 2024{\natexlab{a}}.

\bibitem[Li et~al.(2024{\natexlab{b}})Li, Zhang, Zhang, Zhang, Li, Li, Ma, and Li]{llavaonevision}
Feng Li, Renrui Zhang, Hao Zhang, Yuanhan Zhang, Bo Li, Wei Li, Zejun Ma, and Chunyuan Li.
\newblock Llava-next-interleave: Tackling multi-image, video, and 3d in large multimodal models.
\newblock \emph{arXiv preprint arXiv:2407.07895}, 2024{\natexlab{b}}.

\bibitem[Li et~al.(2022)Li, Li, Xiong, and Hoi]{li2022blip}
Junnan Li, Dongxu Li, Caiming Xiong, and Steven Hoi.
\newblock Blip: Bootstrapping language-image pre-training for unified vision-language understanding and generation.
\newblock In \emph{International conference on machine learning}, pages 12888--12900. PMLR, 2022.

\bibitem[Li et~al.(2023{\natexlab{a}})Li, Li, Savarese, and Hoi]{blip}
Junnan Li, Dongxu Li, Silvio Savarese, and Steven Hoi.
\newblock Blip-2: Bootstrapping language-image pre-training with frozen image encoders and large language models.
\newblock In \emph{International conference on machine learning}, pages 19730--19742. PMLR, 2023{\natexlab{a}}.

\bibitem[Li et~al.(2024{\natexlab{c}})Li, Yuan, Liu, Tang, Wang, Zhu, and Zhang]{tokenpacker}
Wentong Li, Yuqian Yuan, Jian Liu, Dongqi Tang, Song Wang, Jianke Zhu, and Lei Zhang.
\newblock Tokenpacker: Efficient visual projector for multimodal llm.
\newblock \emph{arXiv preprint arXiv:2407.02392}, 2024{\natexlab{c}}.

\bibitem[Li et~al.(2023{\natexlab{b}})Li, Du, Zhou, Wang, Zhao, and Wen]{pope}
Yifan Li, Yifan Du, Kun Zhou, Jinpeng Wang, Wayne~Xin Zhao, and Ji-Rong Wen.
\newblock Evaluating object hallucination in large vision-language models.
\newblock In \emph{The 2023 Conference on Empirical Methods in Natural Language Processing}, 2023{\natexlab{b}}.

\bibitem[Li et~al.(2024{\natexlab{d}})Li, Zhang, Wang, Zhong, Chen, Chu, Liu, and Jia]{minigemini}
Yanwei Li, Yuechen Zhang, Chengyao Wang, Zhisheng Zhong, Yixin Chen, Ruihang Chu, Shaoteng Liu, and Jiaya Jia.
\newblock Mini-gemini: Mining the potential of multi-modality vision language models.
\newblock \emph{arXiv preprint arXiv:2403.18814}, 2024{\natexlab{d}}.

\bibitem[Li et~al.(2024{\natexlab{e}})Li, Yang, Liu, Ma, Zhang, Yang, Sun, Liu, and Bai]{monkey}
Zhang Li, Biao Yang, Qiang Liu, Zhiyin Ma, Shuo Zhang, Jingxu Yang, Yabo Sun, Yuliang Liu, and Xiang Bai.
\newblock Monkey: Image resolution and text label are important things for large multi-modal models.
\newblock In \emph{Proceedings of the IEEE/CVF Conference on Computer Vision and Pattern Recognition}, pages 26763--26773, 2024{\natexlab{e}}.

\bibitem[Liu et~al.(2024{\natexlab{a}})Liu, Li, Li, and Lee]{llava1.5}
Haotian Liu, Chunyuan Li, Yuheng Li, and Yong~Jae Lee.
\newblock Improved baselines with visual instruction tuning.
\newblock In \emph{Proceedings of the IEEE/CVF Conference on Computer Vision and Pattern Recognition}, pages 26296--26306, 2024{\natexlab{a}}.

\bibitem[Liu et~al.(2024{\natexlab{b}})Liu, Li, Li, Li, Zhang, Shen, and Lee]{llavanext}
Haotian Liu, Chunyuan Li, Yuheng Li, Bo Li, Yuanhan Zhang, Sheng Shen, and Yong~Jae Lee.
\newblock Llava-next: Improved reasoning, ocr, and world knowledge, 2024{\natexlab{b}}.

\bibitem[Liu et~al.(2024{\natexlab{c}})Liu, Li, Wu, and Lee]{llava}
Haotian Liu, Chunyuan Li, Qingyang Wu, and Yong~Jae Lee.
\newblock Visual instruction tuning.
\newblock \emph{Advances in neural information processing systems}, 36, 2024{\natexlab{c}}.

\bibitem[Liu et~al.(2024{\natexlab{d}})Liu, Duan, Zhang, Li, Zhang, Zhao, Yuan, Wang, He, Liu, Chen, and Lin]{mmb}
Yuan Liu, Haodong Duan, Yuanhan Zhang, Bo Li, Songyang Zhang, Wangbo Zhao, Yike Yuan, Jiaqi Wang, Conghui He, Ziwei Liu, Kai Chen, and Dahua Lin.
\newblock Mmbench: Is your multi-modal model an all-around player?, 2024{\natexlab{d}}.

\bibitem[Loshchilov et~al.(2017)Loshchilov, Hutter, et~al.]{adam}
Ilya Loshchilov, Frank Hutter, et~al.
\newblock Fixing weight decay regularization in adam.
\newblock \emph{arXiv preprint arXiv:1711.05101}, 5, 2017.

\bibitem[Lu et~al.(2024)Lu, Liu, Zhang, Wang, Dong, Liu, Sun, Ren, Li, Yang, et~al.]{lu2024deepseek}
Haoyu Lu, Wen Liu, Bo Zhang, Bingxuan Wang, Kai Dong, Bo Liu, Jingxiang Sun, Tongzheng Ren, Zhuoshu Li, Hao Yang, et~al.
\newblock Deepseek-vl: towards real-world vision-language understanding.
\newblock \emph{arXiv preprint arXiv:2403.05525}, 2024.

\bibitem[Lu et~al.(2022)Lu, Mishra, Xia, Qiu, Chang, Zhu, Tafjord, Clark, and Kalyan]{sqa}
Pan Lu, Swaroop Mishra, Tanglin Xia, Liang Qiu, Kai-Wei Chang, Song-Chun Zhu, Oyvind Tafjord, Peter Clark, and Ashwin Kalyan.
\newblock Learn to explain: Multimodal reasoning via thought chains for science question answering.
\newblock \emph{Advances in Neural Information Processing Systems}, 35:\penalty0 2507--2521, 2022.

\bibitem[Lu et~al.(2023)Lu, Bansal, Xia, Liu, Li, Hajishirzi, Cheng, Chang, Galley, and Gao]{mathvista}
Pan Lu, Hritik Bansal, Tony Xia, Jiacheng Liu, Chunyuan Li, Hannaneh Hajishirzi, Hao Cheng, Kai-Wei Chang, Michel Galley, and Jianfeng Gao.
\newblock Mathvista: Evaluating mathematical reasoning of foundation models in visual contexts.
\newblock \emph{arXiv preprint arXiv:2310.02255}, 2023.

\bibitem[Luo et~al.(2022)Luo, Zhou, Sun, Wang, Cao, Wu, Huang, and Ji]{luo3}
Gen Luo, Yiyi Zhou, Xiaoshuai Sun, Yan Wang, Liujuan Cao, Yongjian Wu, Feiyue Huang, and Rongrong Ji.
\newblock Towards lightweight transformer via group-wise transformation for vision-and-language tasks.
\newblock \emph{IEEE Transactions on Image Processing}, 31:\penalty0 3386--3398, 2022.

\bibitem[Luo et~al.(2023)Luo, Zhou, Ren, Chen, Sun, and Ji]{lavin}
Gen Luo, Yiyi Zhou, Tianhe Ren, Shengxin Chen, Xiaoshuai Sun, and Rongrong Ji.
\newblock Cheap and quick: Efficient vision-language instruction tuning for large language models.
\newblock In \emph{Advances in Neural Information Processing Systems}, pages 29615--29627. Curran Associates, Inc., 2023.

\bibitem[Luo et~al.(2024{\natexlab{a}})Luo, Zhou, Huang, Ren, Sun, and Ji]{luo1}
Gen Luo, Yiyi Zhou, Minglang Huang, Tianhe Ren, Xiaoshuai Sun, and Rongrong Ji.
\newblock Moil: Momentum imitation learning for efficient vision-language adaptation.
\newblock \emph{IEEE Transactions on Pattern Analysis and Machine Intelligence}, 2024{\natexlab{a}}.

\bibitem[Luo et~al.(2024{\natexlab{b}})Luo, Zhou, Sun, Wu, Gao, and Ji]{luo2}
Gen Luo, Yiyi Zhou, Xiaoshuai Sun, Yongjian Wu, Yue Gao, and Rongrong Ji.
\newblock Towards language-guided visual recognition via dynamic convolutions.
\newblock \emph{International Journal of Computer Vision}, 132\penalty0 (1):\penalty0 1--19, 2024{\natexlab{b}}.

\bibitem[Luo et~al.(2024{\natexlab{c}})Luo, Zhou, Zhang, Zheng, Sun, and Ji]{llavahr}
Gen Luo, Yiyi Zhou, Yuxin Zhang, Xiawu Zheng, Xiaoshuai Sun, and Rongrong Ji.
\newblock Feast your eyes: Mixture-of-resolution adaptation for multimodal large language models.
\newblock \emph{arXiv preprint arXiv:2403.03003}, 2024{\natexlab{c}}.

\bibitem[Masry et~al.(2022)Masry, Long, Tan, Joty, and Hoque]{chartqa}
Ahmed Masry, Do~Xuan Long, Jia~Qing Tan, Shafiq Joty, and Enamul Hoque.
\newblock Chartqa: A benchmark for question answering about charts with visual and logical reasoning, 2022.

\bibitem[Mathew et~al.(2021)Mathew, Karatzas, and Jawahar]{docvqa}
Minesh Mathew, Dimosthenis Karatzas, and C.~V. Jawahar.
\newblock Docvqa: A dataset for vqa on document images, 2021.

\bibitem[Oquab et~al.(2023)Oquab, Darcet, Moutakanni, Vo, Szafraniec, Khalidov, Fernandez, Haziza, Massa, El-Nouby, et~al.]{oquab2023dinov2}
Maxime Oquab, Timoth{\'e}e Darcet, Th{\'e}o Moutakanni, Huy Vo, Marc Szafraniec, Vasil Khalidov, Pierre Fernandez, Daniel Haziza, Francisco Massa, Alaaeldin El-Nouby, et~al.
\newblock Dinov2: Learning robust visual features without supervision.
\newblock \emph{arXiv preprint arXiv:2304.07193}, 2023.

\bibitem[Radford et~al.(2021{\natexlab{a}})Radford, Kim, Hallacy, Ramesh, Goh, Agarwal, Sastry, Askell, Mishkin, Clark, Krueger, and Sutskever]{clip}
Alec Radford, Jong~Wook Kim, Chris Hallacy, Aditya Ramesh, Gabriel Goh, Sandhini Agarwal, Girish Sastry, Amanda Askell, Pamela Mishkin, Jack Clark, Gretchen Krueger, and Ilya Sutskever.
\newblock Learning transferable visual models from natural language supervision, 2021{\natexlab{a}}.

\bibitem[Radford et~al.(2021{\natexlab{b}})Radford, Kim, Hallacy, Ramesh, Goh, Agarwal, Sastry, Askell, Mishkin, Clark, et~al.]{radford2021learning}
Alec Radford, Jong~Wook Kim, Chris Hallacy, Aditya Ramesh, Gabriel Goh, Sandhini Agarwal, Girish Sastry, Amanda Askell, Pamela Mishkin, Jack Clark, et~al.
\newblock Learning transferable visual models from natural language supervision.
\newblock In \emph{International conference on machine learning}, pages 8748--8763. PMLR, 2021{\natexlab{b}}.

\bibitem[Shao et~al.(2024)Shao, Yu, Yu, Ouyang, Zheng, Gai, Wang, and Ding]{shao2024imp}
Zhenwei Shao, Zhou Yu, Jun Yu, Xuecheng Ouyang, Lihao Zheng, Zhenbiao Gai, Mingyang Wang, and Jiajun Ding.
\newblock Imp: Highly capable large multimodal models for mobile devices.
\newblock \emph{arXiv preprint arXiv:2405.12107}, 2024.

\bibitem[Shi et~al.(2024{\natexlab{a}})Shi, Wu, Mao, Wang, and Darrell]{need}
Baifeng Shi, Ziyang Wu, Maolin Mao, Xin Wang, and Trevor Darrell.
\newblock When do we not need larger vision models?, 2024{\natexlab{a}}.

\bibitem[Shi et~al.(2024{\natexlab{b}})Shi, Liu, Wang, Liao, Radhakrishnan, Huang, Yin, Sapra, Yacoob, Shi, Catanzaro, Tao, Kautz, Yu, and Liu]{eagle}
Min Shi, Fuxiao Liu, Shihao Wang, Shijia Liao, Subhashree Radhakrishnan, De-An Huang, Hongxu Yin, Karan Sapra, Yaser Yacoob, Humphrey Shi, Bryan Catanzaro, Andrew Tao, Jan Kautz, Zhiding Yu, and Guilin Liu.
\newblock Eagle: Exploring the design space for multimodal llms with mixture of encoders.
\newblock \emph{arXiv:2408.15998}, 2024{\natexlab{b}}.

\bibitem[Singh et~al.(2019)Singh, Natarajan, Shah, Jiang, Chen, Batra, Parikh, and Rohrbach]{textvqa}
Amanpreet Singh, Vivek Natarajan, Meet Shah, Yu Jiang, Xinlei Chen, Dhruv Batra, Devi Parikh, and Marcus Rohrbach.
\newblock Towards vqa models that can read.
\newblock In \emph{Proceedings of the IEEE/CVF conference on computer vision and pattern recognition}, pages 8317--8326, 2019.

\bibitem[Team et~al.(2024)Team, Mesnard, Hardin, Dadashi, Bhupatiraju, Pathak, Sifre, Rivi{\`e}re, Kale, Love, et~al.]{team2024gemma}
Gemma Team, Thomas Mesnard, Cassidy Hardin, Robert Dadashi, Surya Bhupatiraju, Shreya Pathak, Laurent Sifre, Morgane Rivi{\`e}re, Mihir~Sanjay Kale, Juliette Love, et~al.
\newblock Gemma: Open models based on gemini research and technology.
\newblock \emph{arXiv preprint arXiv:2403.08295}, 2024.

\bibitem[Touvron et~al.(2023)Touvron, Martin, Stone, Albert, Almahairi, Babaei, Bashlykov, Batra, Bhargava, Bhosale, et~al.]{llama}
Hugo Touvron, Louis Martin, Kevin Stone, Peter Albert, Amjad Almahairi, Yasmine Babaei, Nikolay Bashlykov, Soumya Batra, Prajjwal Bhargava, Shruti Bhosale, et~al.
\newblock Llama 2: Open foundation and fine-tuned chat models.
\newblock \emph{arXiv preprint arXiv:2307.09288}, 2023.

\bibitem[Turc et~al.(2019)Turc, Chang, Lee, and Toutanova]{bert}
Iulia Turc, Ming-Wei Chang, Kenton Lee, and Kristina Toutanova.
\newblock Well-read students learn better: On the importance of pre-training compact models.
\newblock \emph{arXiv preprint arXiv:1908.08962v2}, 2019.

\bibitem[Wang et~al.(2024)Wang, Bai, Tan, Wang, Fan, Bai, Chen, Liu, Wang, Ge, et~al.]{wang2024qwen2}
Peng Wang, Shuai Bai, Sinan Tan, Shijie Wang, Zhihao Fan, Jinze Bai, Keqin Chen, Xuejing Liu, Jialin Wang, Wenbin Ge, et~al.
\newblock Qwen2-vl: Enhancing vision-language model's perception of the world at any resolution.
\newblock \emph{arXiv preprint arXiv:2409.12191}, 2024.

\bibitem[Xu(2015)]{show}
Kelvin Xu.
\newblock Show, attend and tell: Neural image caption generation with visual attention.
\newblock \emph{arXiv preprint arXiv:1502.03044}, 2015.

\bibitem[Yang et~al.(2016)Yang, He, Gao, Deng, and Smola]{stacked}
Zichao Yang, Xiaodong He, Jianfeng Gao, Li Deng, and Alex Smola.
\newblock Stacked attention networks for image question answering.
\newblock In \emph{Proceedings of the IEEE conference on computer vision and pattern recognition}, pages 21--29, 2016.

\bibitem[Ye et~al.(2024{\natexlab{a}})Ye, Xu, Xu, Ye, Yan, Zhou, Wang, Hu, Shi, Shi, Li, Xu, Chen, Tian, Qian, Zhang, Huang, and Zhou]{mplug}
Qinghao Ye, Haiyang Xu, Guohai Xu, Jiabo Ye, Ming Yan, Yiyang Zhou, Junyang Wang, Anwen Hu, Pengcheng Shi, Yaya Shi, Chenliang Li, Yuanhong Xu, Hehong Chen, Junfeng Tian, Qi Qian, Ji Zhang, Fei Huang, and Jingren Zhou.
\newblock mplug-owl: Modularization empowers large language models with multimodality, 2024{\natexlab{a}}.

\bibitem[Ye et~al.(2024{\natexlab{b}})Ye, Wu, Lin, and Zhou]{ye2024fit}
Weihao Ye, Qiong Wu, Wenhao Lin, and Yiyi Zhou.
\newblock Fit and prune: Fast and training-free visual token pruning for multi-modal large language models.
\newblock \emph{arXiv preprint arXiv:2409.10197}, 2024{\natexlab{b}}.

\bibitem[Yu et~al.(2023)Yu, Yang, Li, Wang, Lin, Liu, Wang, and Wang]{mmvet}
Weihao Yu, Zhengyuan Yang, Linjie Li, Jianfeng Wang, Kevin Lin, Zicheng Liu, Xinchao Wang, and Lijuan Wang.
\newblock Mm-vet: Evaluating large multimodal models for integrated capabilities, 2023.

\bibitem[Yu et~al.(2019)Yu, Yu, Cui, Tao, and Tian]{deep}
Zhou Yu, Jun Yu, Yuhao Cui, Dacheng Tao, and Qi Tian.
\newblock Deep modular co-attention networks for visual question answering.
\newblock In \emph{Proceedings of the IEEE/CVF conference on computer vision and pattern recognition}, pages 6281--6290, 2019.

\bibitem[Yuan et~al.(2024)Yuan, Li, Huang, Ye, and Sun]{yuan2024tinygptvefficientmultimodallarge}
Zhengqing Yuan, Zhaoxu Li, Weiran Huang, Yanfang Ye, and Lichao Sun.
\newblock Tinygpt-v: Efficient multimodal large language model via small backbones, 2024.

\bibitem[Yue et~al.(2024)Yue, Ni, Zhang, Zheng, Liu, Zhang, Stevens, Jiang, Ren, Sun, Wei, Yu, Yuan, Sun, Yin, Zheng, Yang, Liu, Huang, Sun, Su, and Chen]{mmmu}
Xiang Yue, Yuansheng Ni, Kai Zhang, Tianyu Zheng, Ruoqi Liu, Ge Zhang, Samuel Stevens, Dongfu Jiang, Weiming Ren, Yuxuan Sun, Cong Wei, Botao Yu, Ruibin Yuan, Renliang Sun, Ming Yin, Boyuan Zheng, Zhenzhu Yang, Yibo Liu, Wenhao Huang, Huan Sun, Yu Su, and Wenhu Chen.
\newblock Mmmu: A massive multi-discipline multimodal understanding and reasoning benchmark for expert agi.
\newblock In \emph{Proceedings of CVPR}, 2024.

\bibitem[Zhai et~al.(2023)Zhai, Mustafa, Kolesnikov, and Beyer]{siglip}
Xiaohua Zhai, Basil Mustafa, Alexander Kolesnikov, and Lucas Beyer.
\newblock Sigmoid loss for language image pre-training, 2023.

\bibitem[Zhang et~al.(2024{\natexlab{a}})Zhang, Gao, Gan, Dufter, Wenzel, Huang, Shah, Du, Zhang, Li, et~al.]{mm1}
Haotian Zhang, Mingfei Gao, Zhe Gan, Philipp Dufter, Nina Wenzel, Forrest Huang, Dhruti Shah, Xianzhi Du, Bowen Zhang, Yanghao Li, et~al.
\newblock Mm1. 5: Methods, analysis \& insights from multimodal llm fine-tuning.
\newblock \emph{arXiv preprint arXiv:2409.20566}, 2024{\natexlab{a}}.

\bibitem[Zhang et~al.(2024{\natexlab{b}})Zhang, Li, Zhang, Pu, Cahyono, Hu, Liu, Zhang, Yang, Li, and Liu]{lmms-eval}
Kaichen Zhang, Bo Li, Peiyuan Zhang, Fanyi Pu, Joshua~Adrian Cahyono, Kairui Hu, Shuai Liu, Yuanhan Zhang, Jingkang Yang, Chunyuan Li, and Ziwei Liu.
\newblock Lmms-eval: Reality check on the evaluation of large multimodal models, 2024{\natexlab{b}}.

\bibitem[Zhang et~al.(2021)Zhang, Li, Hu, Yang, Zhang, Wang, Choi, and Gao]{vinvl}
Pengchuan Zhang, Xiujun Li, Xiaowei Hu, Jianwei Yang, Lei Zhang, Lijuan Wang, Yejin Choi, and Jianfeng Gao.
\newblock Vinvl: Revisiting visual representations in vision-language models.
\newblock In \emph{Proceedings of the IEEE/CVF conference on computer vision and pattern recognition}, pages 5579--5588, 2021.

\bibitem[Zhang et~al.(2022)Zhang, Roller, Goyal, Artetxe, Chen, Chen, Dewan, Diab, Li, Lin, et~al.]{opt}
Susan Zhang, Stephen Roller, Naman Goyal, Mikel Artetxe, Moya Chen, Shuohui Chen, Christopher Dewan, Mona Diab, Xian Li, Xi~Victoria Lin, et~al.
\newblock Opt: Open pre-trained transformer language models.
\newblock \emph{arXiv preprint arXiv:2205.01068}, 2022.

\bibitem[Zhou et~al.(2019)Zhou, Ji, Su, Li, and Sun]{free}
Yiyi Zhou, Rongrong Ji, Jinsong Su, Xiangming Li, and Xiaoshuai Sun.
\newblock Free vqa models from knowledge inertia by pairwise inconformity learning.
\newblock In \emph{Proceedings of the AAAI Conference on Artificial Intelligence}, pages 9316--9323, 2019.

\bibitem[Zhou et~al.(2021)Zhou, Ren, Zhu, Sun, Liu, Ding, Xu, and Ji]{trar}
Yiyi Zhou, Tianhe Ren, Chaoyang Zhu, Xiaoshuai Sun, Jianzhuang Liu, Xinghao Ding, Mingliang Xu, and Rongrong Ji.
\newblock Trar: Routing the attention spans in transformer for visual question answering.
\newblock In \emph{Proceedings of the IEEE/CVF International Conference on Computer Vision (ICCV)}, pages 2074--2084, 2021.

\bibitem[Zhu et~al.(2023)Zhu, Chen, Shen, Li, and Elhoseiny]{minigpt}
Deyao Zhu, Jun Chen, Xiaoqian Shen, Xiang Li, and Mohamed Elhoseiny.
\newblock Minigpt-4: Enhancing vision-language understanding with advanced large language models, 2023.

\bibitem[Zhu et~al.(2024)Zhu, Zhu, Liu, Liu, Xu, Shen, Peng, Ou, Feng, and Tang]{llavaphi}
Minjie Zhu, Yichen Zhu, Xin Liu, Ning Liu, Zhiyuan Xu, Chaomin Shen, Yaxin Peng, Zhicai Ou, Feifei Feng, and Jian Tang.
\newblock A comprehensive overhaul of multimodal assistant with small language models.
\newblock \emph{arXiv preprint arXiv:2403.06199}, 2024.

\end{thebibliography}
}
\clearpage
\appendix
\section{Quantitative analysis}
\subsection{Impact of EmbQ configuration }
Table~\ref{suppl1} examines the impact of EmbQ's dimensions, number of layers, and insertion positions within the LLM on performance. The first and second blocks show that EmbQ achieves optimal results with a simple configuration of 576 dimensions and a single layer, aligning with FlashSloth's efficiency goals. The third block explores the effects of inserting EmbQ into shallow, middle, deep, or multiple layers. Excessively shallow insertions limit Query Tokens to self-interaction, restricting their ability to capture rich image and text features, while overly deep insertions hinder effective propagation of newly learned supplementary information to generated tokens.
\begin{table}[h]
\centering
\caption{Ablation study on the EmbQ configuration, analyzing the effects of the EmbQ dimensions, the number of EmbQ layers, and the EmbQ insertion layers. The experiments are conducted on four benchmarks. The configuration marked with \ddag represents the final selected setting.
}
\renewcommand{\arraystretch}{0.85}
\begin{tabular}{@{}ccccc@{}}
\toprule
\multicolumn{1}{c|}{\textbf{Choices}} & \textbf{GQA} & \textbf{POPE} & \textbf{MME} & \textbf{MMB} \\ \midrule
\multicolumn{5}{c}{Dimension of EmbQ}                                                              \\ \midrule
\multicolumn{1}{c|}{576 \ddag}              & 60.8         & 87.8          & 1490.7       & 67.9         \\
\multicolumn{1}{c|}{768}              & 60.6         & 86.8          & 1446.7       & 66.6         \\
\multicolumn{1}{c|}{1152}             & 60.6         & 86.8          & 1455.5       & 66.4         \\
\multicolumn{1}{c|}{2560}             & 60.5         & 87.1          & 1420.8       & 67.1         \\ \midrule
\multicolumn{5}{c}{Number of EmbQ layer}                                                           \\ \midrule
\multicolumn{1}{c|}{1 \ddag}                & 60.8         & 87.8          & 1490.7       & 67.9         \\
\multicolumn{1}{c|}{2}                & 60.6         & 86.9          & 1488.7       & 67.1         \\
\multicolumn{1}{c|}{3}                & 60.6         & 87.2          & 1457.5       & 67.3         \\ \midrule
\multicolumn{5}{c}{EmbQ Insertion Layer}                                                           \\ \midrule
\multicolumn{1}{c|}{4}                & 60.7         & 86.9          & 1433.3       & 66.2         \\
\multicolumn{1}{c|}{8 \ddag}                & 60.8         & 87.8          & 1490.7       & 67.9         \\
\multicolumn{1}{c|}{16}               & 60.4         & 87.3          & 1399.2       & 66.5         \\
\multicolumn{1}{c|}{24}               & 60.8         & 87.1          & 1433.7       & 67.8         \\
\multicolumn{1}{c|}{8/16/24}          & 60.8         & 86.1          & 1414.5       & 67.0         \\ \bottomrule
\end{tabular}
\label{suppl1}
\end{table}
\subsection{Impact of Feature Fusion Methods}
Table~\ref{suppl_2} compares methods for fusing features learned by EmbQ with Query Token features: direct replacement, addition, and gated fusion~\cite{llavahr}, which adjusts the contribution of each feature. Results show that direct addition achieves the best performance by effectively integrating EmbQ's features while retaining the original feature information.
\begin{table}[]
\centering
\caption{Ablation experiment results on four benchmarks under different feature fusion method}
\begin{tabular}{@{}c|cccc@{}}
\toprule
\textbf{Feature Fusion} & \textbf{GQA} & \textbf{POPE} & \textbf{MME} & \textbf{MMB} \\ \midrule
Add                     & \textbf{60.8}         & \textbf{87.8}          & \textbf{1490.7}       & \textbf{67.9}         \\
Replace                 & 60.5         & 86.3          & 1443.7       & 66.4         \\
Gate fusion             & 60.7         & 86.1          & 1468.1       & 66.1         \\ \bottomrule
\end{tabular}
\label{suppl_2}
\end{table}
\subsection{Comparison with Other Vision Compression Methods}
\begin{table}[t]
\centering
\caption{Experimental results with EmbQ combined with various vision compression method to compress visual tokens to 81.}
\begin{tabular}{@{}l|cccc@{}}
\toprule
\textbf{Method}                                                                & \textbf{GQA} & \textbf{MME} & \textbf{MMB} & \textbf{POPE} \\ \midrule
\multirow{2}{*}{\begin{tabular}[c]{@{}l@{}}Avg. pooling\\ +EmbQ\end{tabular}}  & 59.6         & 1440.3       & 62.7         & 84.4          \\
                                                                               & 59.8         & 1454.7       & 66.4         & 86.4          \\ \midrule
\multirow{2}{*}{\begin{tabular}[c]{@{}l@{}}Att. pooling\\ +EmbQ\end{tabular}}  & 60.3         & 1444.8       & 63.9         & 86.4          \\
                                                                               & \textbf{60.8}         & \textbf{1490.7 }      & \textbf{67.9}         & \textbf{87.8}          \\ \midrule
\multirow{2}{*}{\begin{tabular}[c]{@{}l@{}}Pixel shuffle~\cite{internvl}\\ +EmbQ\end{tabular}} & 58.6         & 1443.5       & 65.8         & 85.0          \\
                                                                               & 59.2         & 1445.3       & 66.0         & 86.1          \\ \midrule
\multirow{2}{*}{\begin{tabular}[c]{@{}l@{}}LDP-V2~\cite{chu2024mobilevlm}\\ +EmbQ\end{tabular}}        & 59.4         & 1438.5       & 65.3         & 85.6          \\
                                                                               & 59.9         & 1464.6       & 66.6         & 85.9          \\ \bottomrule
\end{tabular}
\label{suppl3}
\end{table}
Table~\ref{suppl3} compares our method with various visual feature compression approaches and evaluates the performance gains from incorporating EmbQ. The combination of SAP and EmbQ achieves the best results, validating the integration of SAP's saliency features with EmbQ's instruction-related features. Additionally, embedding EmbQ into any compression method consistently improves performance, highlighting its effectiveness.

\section{Qualitative Analysis}
This section showcases practical examples of FlashSloth and FlashSloth-HD, demonstrating their real-world performance in multidisciplinary question answering, code generation, real-world scene reasoning, fine-grained text information extraction, and chart analysis reasoning. FlashSloth delivers accurate responses and exceptional performance across these tasks.

\begin{figure*}
    \centering
    \includegraphics[width=0.9\textwidth]{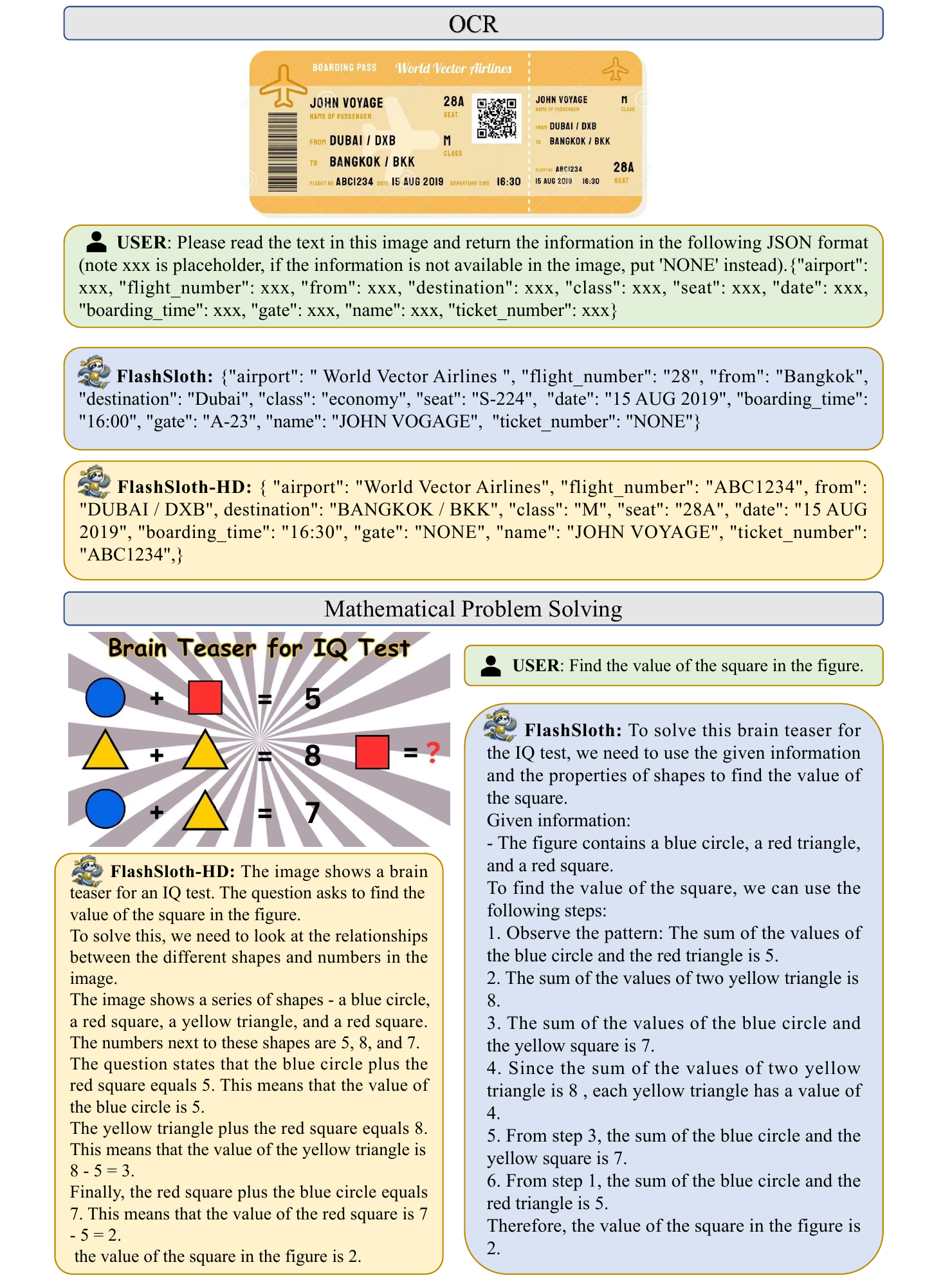}
    \caption{FlashSloth's Performance in Ticket OCR Recognition and Mathematical Question Answering.}
\end{figure*}
\begin{figure*}
    \centering
    \includegraphics[width=0.9\textwidth]{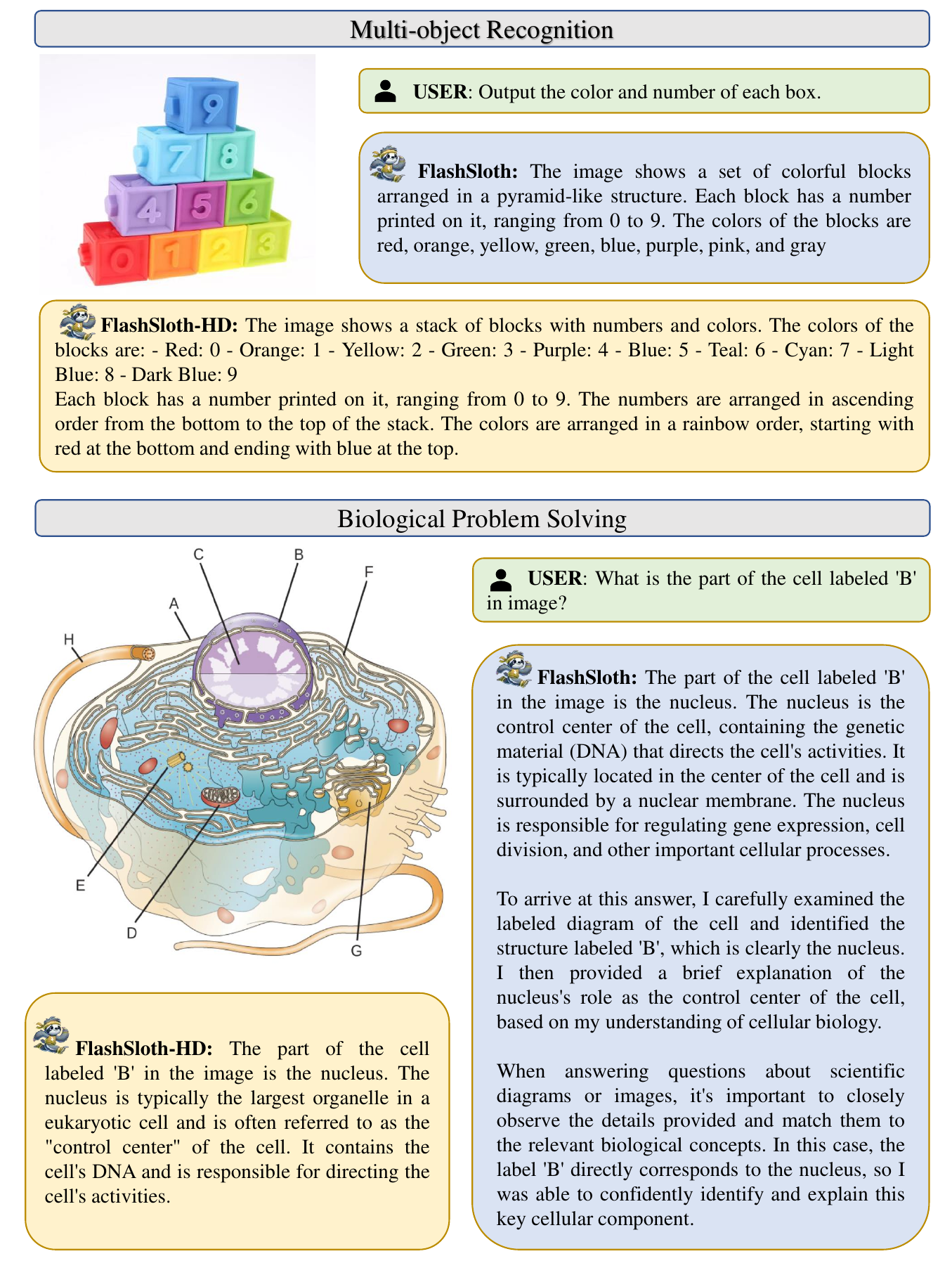}
    \caption{FlashSloth's Performance in Recognizing the Colors and Content of Multiple Objects and on Biological Question Answering Tasks.}
\end{figure*}
\begin{figure*}
    \centering
    \includegraphics[width=0.9\textwidth]{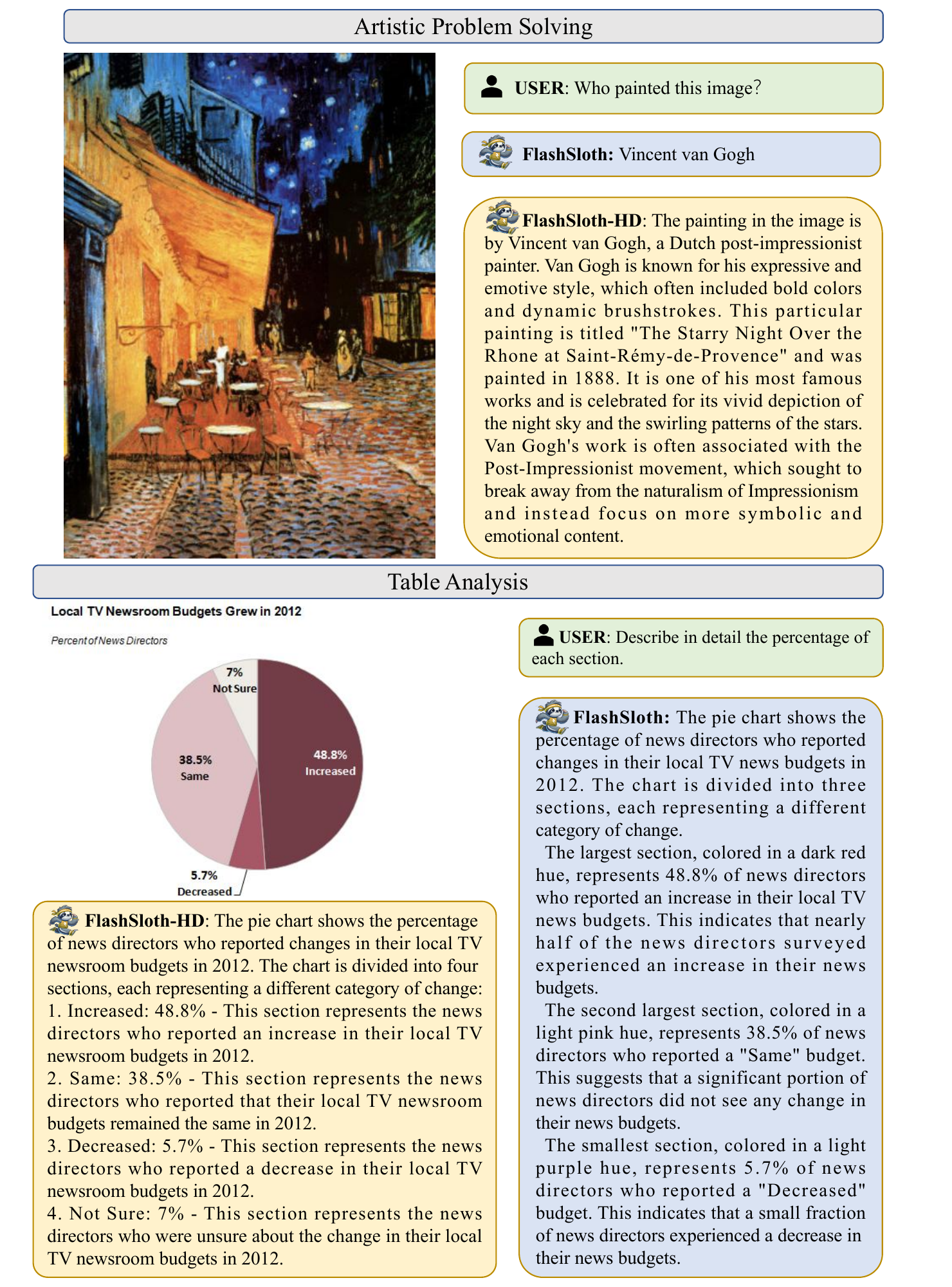}
    \caption{FlashSloth's Performance in Artistic Question Answering and Chart Content Analysis.}
\end{figure*}
\begin{figure*}
    \centering
    \includegraphics[width=0.9\textwidth]{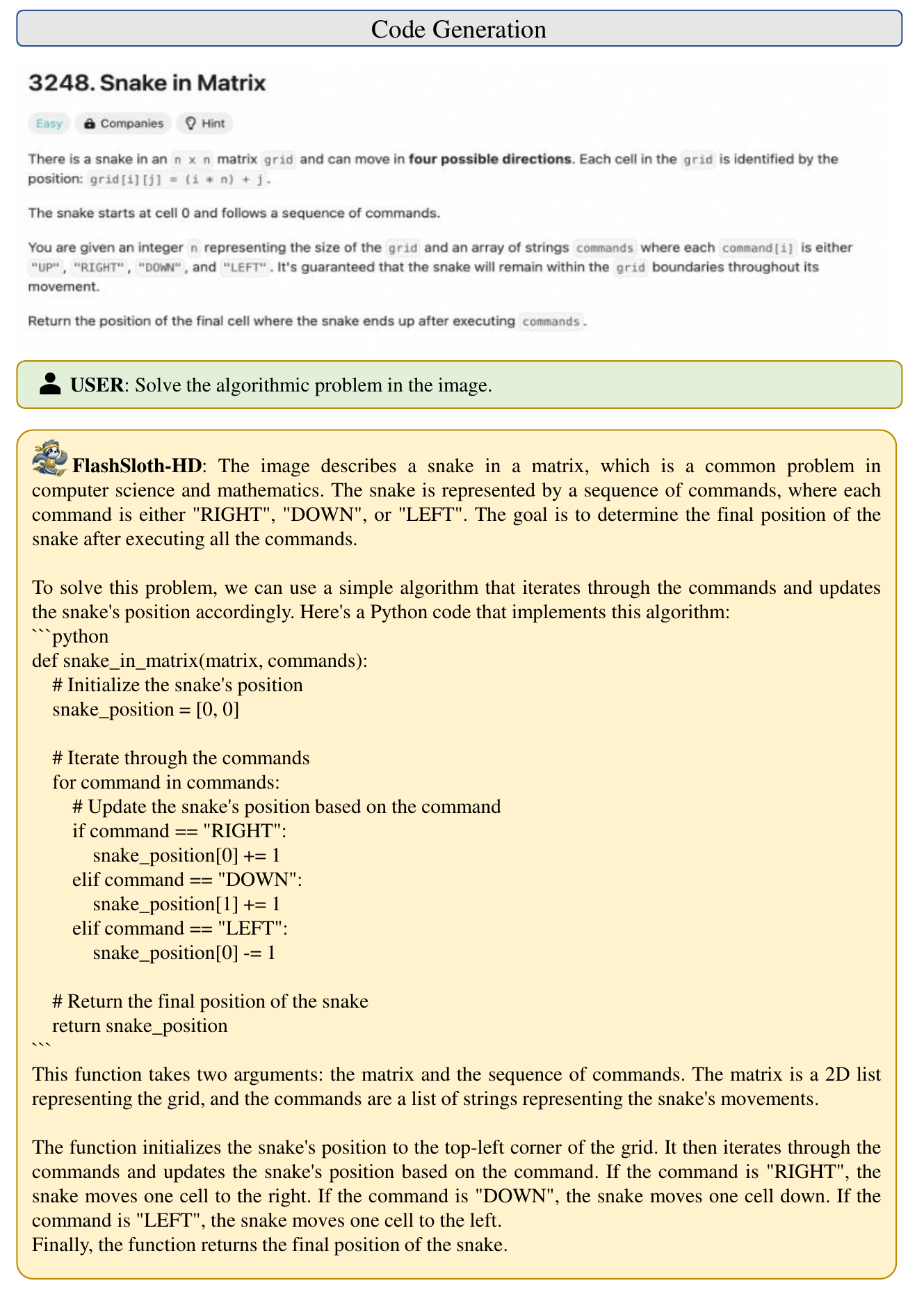}
    \caption{FlashSloth's Performance in Solving Algorithmic Problems Using Code.}
\end{figure*}
\begin{figure*}[]
    \centering
    \includegraphics[width=0.89\textwidth]{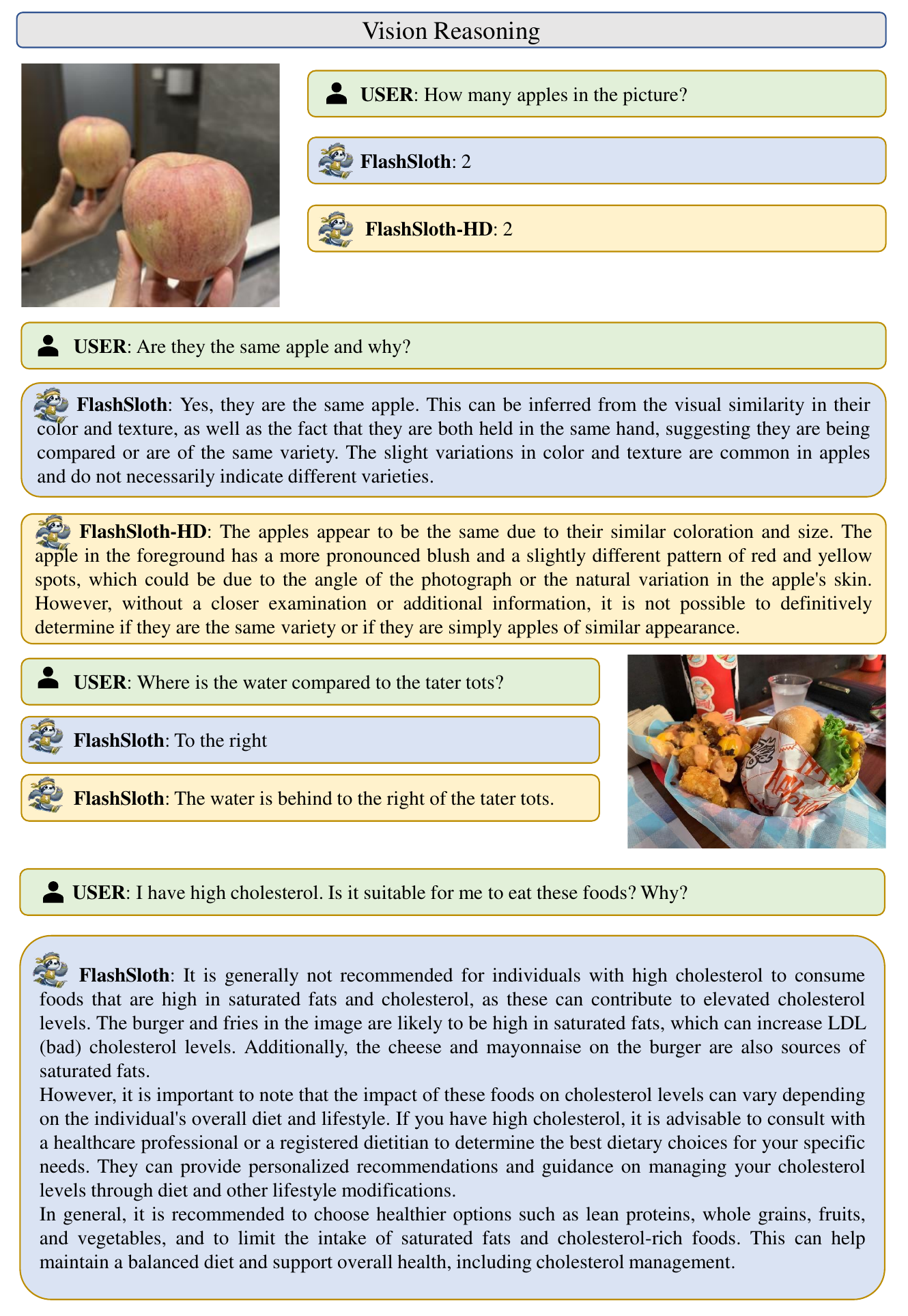}
    \caption{FlashSloth's Performance in Real-World Scene Reasoning Tasks.}
\end{figure*}

\clearpage

\end{document}